\documentclass[aps,pre,9pt,twocolumn,superscriptaddress,floatfix]{revtex4-2}

\usepackage{microtype}
\usepackage{graphicx}
\usepackage{subfigure}
\usepackage{nicematrix}
\usepackage{booktabs} 
\PassOptionsToPackage{hyphens}{url}
\usepackage{hyperref}

\newcommand{\n}[1]{\| #1 \|}

\usepackage{amsmath}
\usepackage{amssymb}
\usepackage{mathtools}
\usepackage{amsthm}
\usepackage{enumitem}

\usepackage{orcidlink}
\usepackage{lipsum, babel}

\usepackage[capitalize,noabbrev]{cleveref}

\theoremstyle{plain}
\newtheorem{theorem}{Theorem}[section]

\newtheorem{lemma}[theorem]{Lemma}

\theoremstyle{definition}

\newtheorem{assumption}[theorem]{Assumption}
\theoremstyle{remark}

\usepackage{multirow}
\usepackage{algorithm2e}
\usepackage{algorithmic}

\usepackage[textsize=tiny]{todonotes}

\def\tilde{\widetilde}
\DeclareMathOperator*{\argmin}{argmin}
\DeclareMathOperator*{\argmax}{argmax}

\newcommand{\RNum}[1]{\uppercase\expandafter{\romannumeral #1\relax}}

\usepackage{xcolor}

\begin{document}

\title{Learning the Right Layers: a Data-Driven Layer-Aggregation Strategy for Semi-Supervised Learning on Multilayer Graphs}

\author{Sara Venturini\,\orcidlink{0000-0002-2653-8533}}
\email[Corresponding author: ]{sara.venturini@math.unipd.it}
\affiliation{Department of Mathematics “Tullio Levi-Civita”, University of Padova, Padova 35121, Italy}

\author{Andrea Cristofari\,\orcidlink{0000-0002-9126-3994}}
\affiliation{Department of Civil Engineering and Computer Science Engineering, University of Rome ``Tor Vergata'', Rome 00133, Italy}

\author{Francesco Rinaldi\,\orcidlink{0000-0001-8978-6027}} 
\affiliation{Department of Mathematics “Tullio Levi-Civita”, University of Padova, Padova 35121, Italy}

\author{Francesco Tudisco\,\orcidlink{0000-0002-8150-4475}}
\affiliation{School of Mathematics, Gran Sasso Science Institute, L'Aquila 67100, Italy}


\begin{abstract}
Clustering  (or community detection) on multilayer graphs poses several additional complications with respect to standard graphs as different layers may be characterized by different structures and types of information. One of the major challenges is to establish the extent to which each layer contributes to the cluster assignment in order to effectively take advantage of the multilayer structure and improve upon the classification obtained using the individual layers or their union. However, making an informed a-priori assessment about the clustering information content of the layers can be very complicated. In this work, we assume a semi-supervised learning setting, where the class of a small percentage of nodes is initially provided,  and we propose a parameter-free Laplacian-regularized model that learns an optimal nonlinear combination of the different layers from the available input labels. The learning algorithm is based on a Frank-Wolfe optimization scheme with inexact gradient, combined with a modified Label Propagation iteration. We provide a detailed convergence analysis of the algorithm and extensive experiments on synthetic and real-world datasets, showing that the proposed method  compares favourably with a variety of baselines and  outperforms each individual layer when used in isolation.
\end{abstract}

\maketitle

\section{Introduction}
\label{Introduction}
Graph-based Semi-Supervised Learning (GSSL) has achieved great success in various real-world applications, where only a relatively small amount of labeled samples are available \cite{song2022graph}. Given a graph and a set of initially labeled nodes, the aim of GSSL is to infer the labels of the remaining unlabeled nodes. To this end, exploiting the graph structure is particularly important, especially when the percentage of input labels (size of the training dataset) is small and when no features are available for the nodes. Based on the so-called smoothness assumption, one of the most successful approaches for GSSL relies on a Laplacian regularization formulation, where we aim at minimizing a loss function that simultaneously forces consistency with the initial labels and with the graph structure.  The minimizer can be interpreted as a new node embedding, which is then used to classify the unlabeled nodes. After the pioneering work by \citet{zhou2003learning,belkin2004regularization,yang2016revisiting}, this approach has been widely explored in the machine learning literature \cite{hein2013total,welling2016semi,gasteiger2018predict,mercado2019generalized,huang2020combining,tudisco2021nonlinear,prokopchik2022nonlinear}.

While graphs are a popular and successful tool to model data interactions, many empirical systems and real-world datasets are characterized by multiple types of interactions or relationships simultaneously, and are actually better described by multilayer graphs \cite{de2013mathematical,gao2012networks,wang2015evolutionary}. 
For instance, transportation systems are characterized by different transportation means such as train, bus, etc \cite{de2014navigability}, 
scientific data is characterized by co-authorship, co-citation, as well as topic and institution affinities \cite{higham2022multilayer},   people
in a social environment interact at different layers such as friendship, acquaintance or business, etc \cite{dickison2016multilayer}. Also, many biological systems are characterized by 
multiple types of relationships among their constituents \cite{bentley2016multilayer,mangioni2018multilayer}.  Multilayer graphs are a standard representation of such data and directly modeling these multilayer interactions has led to improvements in a number of network science and machine learning problems~\cite{bianconi2018multilayer,magnani2021community}.

Even though potentially very useful and powerful, multilayer graph models pose an intrinsic fundamental challenge. Multilayer graphs can have a large number of layers describing a variety of different properties, however,  it is a-priori not clear whether all the layers are actually useful to classify the nodes. Layers may carry the same or complementary clustering information,  some layers may be more informative than others, and certain layers can potentially be just noise (i.e.\ they carry no information about the node clusters). 
Deciding which of these situations better describes a given dataset and identifying which are the most (and the least) informative layers is both very useful and highly challenging. In fact, the construction of the networks in many applications is not straightforward, and making an informed a-priori assessment on the presence of noise, the different types of layer structures, and the clustering information content in general, can be very complicated \cite{bentley2016multilayer,choobdar2019assessment,peel2022statistical}.

Several GSSL algorithms for multilayer networks have been developed in recent years. The majority of these methods propose to aggregate the information carried by the different layers into a single-layer graph, using different forms of aggregation functions, such as sum, min, max, etc  
\cite{tsuda2005fast,argyriou2005combining,zhou2007spectral,kato2008robust,karasuyama2013multiple,nie2016parameter,mondragon2018multilink,ye2018robust,mercado2019generalized,viswanathan2019improved,bergermann2021semi}. 
However, most of the time, the proposed combinations are meant to be effective for a particular setting and thus require a-priori knowledge of the type of clustering information the layers and the whole dataset carry along. Moreover, even though some methods perform  well in more than one setting, e.g.\ \cite{mercado2019generalized,gujral2018smacd}, they do not provide information on whether certain layers are more informative than others, whether the information is complementary or not,  or whether there is some uninformative (noisy) layer.

In this work, we propose a parameter-free Laplacian-regularized model that learns an optimal combination of the different layers from the available input labels. In our model, the layers are combined via nonlinear generalized mean functions which include as special cases several aggregation functions previously used in the literature. The optimal aggregation parameters are computed via a tailored bilevel inexact-gradient optimization scheme. We provide a detailed convergence analysis of the optimization method, which we also extensively test numerically. Our  tests on synthetic and real-world datasets show that the resulting  GSSL method for multilayer networks very favorably compares with available alternatives in terms of accuracy performance and that it succeeds in identifying the most relevant and least relevant layers, as well as complementary information across the layers.

\section{Related Work on Multilayer GSSL}
\label{Related Work}

In this section, we provide a brief review of available semi-supervised learning algorithms for multilayer graphs, mostly focusing  on those designed to work on feature-less multilayer networks. 

Similar to our model, several approaches are based on  learning  an optimal set of parameters in the aggregation function of the multilayer graph, leveraging a multilayer version of  the  Laplacian-regularization  formulation of GSSL.
\citet{tsuda2005fast} propose a method for protein classification using multiple protein networks. The multilayer graph is aggregated via a weighted linear  combination, whose weights are learned in a variational min-max fashion aimed at minimizing the worst-case graph consistency function. The resulting method  performs well in the presence of noisy or irrelevant layers. Similarly, 
\citet{argyriou2005combining}  compute an optimal linear combination of Laplacian kernels, which solves an extended regularization problem on the multilayer graph, enforcing a joint minimization over both the data and the set of graph kernels. 
Then, \citet{zhou2007spectral} show that the resulting convex combination of graph Laplacians generalizes the  normalized cut function to multilayer networks. 
An alternative formulation is proposed in \cite{nie2016parameter} where the optimal weights in the weighted linear combination of layers' Laplacian are defined implicitly via a dual Lagrangian formulation. The resulting method is  a parameter-free method for optimal layer weights.  
Finally, \citet{karasuyama2013multiple} suggest an approach to   efficiently linearly aggregate  multiple graphs under the Laplacian regularization framework, by performing a form of alternate optimization via label propagation combined with sparse integration. 
Unlike the approach we propose here, all these methods are based on linear aggregation functions (convex combinations) and the optimal weights are model-based, rather than data-driven, aiming at optimizing some form of worst-case setting.

Nonlinear layer aggregation functions such as max, min, and their generalization, provide additional modeling power. Using a Log-Euclidean matrix function formulation of the generalized power mean of graph Laplacians, \citet{mercado2019generalized} propose a regularizer based on  a one-parameter family of matrix means that includes the arithmetic, geometric and harmonic means as particular cases.
This approach is revised and improved in \cite{bergermann2021semi}, based on diffuse interface methods and fast matrix-vector products.  While able to reach competitive performance, this approach requires an extensive exploration of the parameter defining the mean, which can be computationally prohibitive. Entrywise minimum and maximum aggregation functions are used in the multilink  model proposed in \cite{mondragon2018multilink,bianconi2018multilayer,ghorbanchian2022hyper}. 

Deviating from the Laplacian regularization formulation, 
\citet{eswaran2017zoobp} propose a method based on fast belief propagation
on heterogeneous graphs, with nodes of different types, while 
\citet{gujral2018smacd} design a parameter-free algorithm based on tensor factorization, which  aims at finding both overlapping and non-overlapping communities. 

While very popular in the single-layer setting, only a few  extensions of geometric deep learning and graph neural networks to the multilayer setting are available so far and are mostly designed for the case of multilayer graphs having intra-layers connections, i.e.\ edges connecting different nodes from one layer to another. 
Among the available ones,  \cite{ghorbani2019mgcn} is based on an extension of the graph convolutional filter by \citet{welling2016semi}, while \cite{grassia2021mgnn} proposes a graph neural network whose graph filter is a parametric aggregated Laplacian,  parametrized in terms of an MLP. 



\section{Learning the Most Relevant Layers}
\label{Theory}

\subsection*{Problem Set-up}
We consider a multilayer graph, specifically a multiplex (alternatively known as multicolor, or multiview graph), defined as $\boldsymbol G=\{G^{(1)},\dots,G^{(K)}\}$, with $K$ layers, each being a weighted  undirected graph $G^{(k)}=(V,E^{(k)}, w_k)$ with $V=\{1,\dots,N\}$, $E^{(k)}\subseteq V\times V$, and $w_k:E_k\to\mathbb R_+$. To each layer corresponds a weighted adjacency matrix $A^{(k)}$, whose entries $A_{ij}^{(k)}=w_k(ij)>0$ represent the strength of the tie between  $i$ and $j$, if $ij\in E$, and $A_{ij}^{(k)}=0$ if $ij\notin E$.

Using the terminology proposed  in \cite{magnani2021community}, we assume $\boldsymbol G$  consists of a set $C=\{C_1,\dots,C_m\}$ of communities (or labels) that is total (i.e., every node belongs to at least one $C_j\in C$), node-disjoint (i.e., no node belongs
to more than one cluster), and pillar (i.e., each node belongs to the same community across the layers). Further, we assume that for each $C_j\in C$ we are given a set of input known labels $O_j\subseteq V$ which  are one-hot encoded into the matrix $Y\in\mathbb{R}^{n\times m}$, with $Y_{ij}=1$ if $i\in C_j$, and $Y_{ij}=0$ otherwise.

The goal is to learn the unknown labels. 
In our setting, we assume no node feature is available. In other words, we focus on the  setting in which one has access only to topological information about the graph structure, and has some input knowledge about the community assignment of some nodes. This is a common setting in e.g. network and social science applications \cite{magnani2021community}.





\subsection*{Generalized Mean Adjacency Model}
In order to learn a classifier that effectively takes into account the multilayer graph structure, we design a nonlinear aggregation strategy that optimally learns the aggregation parameters and computes a classifier based on a  multilayer Laplacian-regularization model. To this end, we first briefly review the standard Laplacian-regularization model for single-layer graphs.

If only one layer is available, i.e.\ if we are dealing with the standard graph case, a successful approach to impose local and global consistency with the available input labels and with the graph structure is to minimize the following Laplacian-regularized GSSL loss function
%
\begin{equation}
\label{eq:single_layer_multi_label_opt_problem}
    \textstyle{\varphi(X) := \|X-Y\|_F^2+\frac\lambda 2 \mathrm{Tr}(X^\top L X)}
\end{equation}
over all $X\in \mathbb R^{n\times m}$. Here $L=D-A$ is the Laplacian matrix of the single-layer graph at hand, with $D=\mathrm{diag}(A\mathbf{1})$ the diagonal matrix of the (weighted) degrees. Simple linear algebra passages show that the obtained solution $X^* = \argmin \varphi(X)$ is entrywise positive, thus the entries $X_{ij}^*$ can be interpreted as a classifier that provides a score quantifying the likelihood that node $i$ belongs to community $C_j$. Hence, we assign to each node $i$ the label $C_{k^*}$, with $k^*=\argmax_j X_{ij}^*$.  Note that, if $y^{(k)}$ is the $k$-th column of $Y$, with one-hot information about the input labels in $O_k$, then one can equivalently write $\varphi(X)=\sum_{k=1}^m \varphi_k(x^{(k)})$, where $x^{(k)}$ are the columns of $X$ and 
\begin{equation} 
\label{eq:single_layer_single_label_opt_problem}\varphi_k(x)=\sum_{i=1}^{N} |x_i-y_i^{(k)}|^2 + \frac{\lambda}{2} \sum_{i,j=1}^{N} A_{ij}
    \left( x_{i} - x_{j} \right)^2\, .
\end{equation}
Clearly, as  the $\varphi_k$ are independent of each other, in this single-graph setting minimizing $\varphi$ is equivalent to minimizing each $\varphi_k$ individually.

When we are given $K$ layers, imposing smoothness with respect to the edge structure is more challenging. As the communities are assumed to be consistent across the layers, a standard approach is to tackle the problem after layer aggregation. If the aggregating function is linear, this  boils down to choosing a set of weights $\boldsymbol{\beta}_k>0$ with $\sum_k \boldsymbol{\beta}_k =1$ and replace $A$ in \eqref{eq:single_layer_multi_label_opt_problem} or \eqref{eq:single_layer_single_label_opt_problem} with $A^{lin} = \sum_k \boldsymbol{\beta}_k A^{(k)}$. This approach has been widely explored and is considered for example in \cite{tsuda2005fast,argyriou2005combining,ye2018robust}. As the cost function in \eqref{eq:single_layer_multi_label_opt_problem} is quadratic, this is equivalent to considering the classifier $X^* = \sum_k \boldsymbol{\beta}_k X_k^*$, with $X_k^*$ solution to \eqref{eq:single_layer_multi_label_opt_problem} for $A=A^{(k)}$. Another possibility is to consider ``nonlinear aggregations'' \cite{mondragon2018multilink, mercado2019generalized,bergermann2021semi,ghorbanchian2022hyper}. 
For example, using the concept of multilinks \cite{bianconi2018multilayer}, \citet{mondragon2018multilink} replace the multilayer network with a single-layer graph  with adjacency matrix with entries $A^{max}_{ij}=\max_{k}A_{ij}^{(k)}$ or $A^{min}_{ij}=\min_{k}A_{ij}^{(k)}$. 
%
%
%
The maximum-based model
corresponds to assuming the edge $ij$ exists in the aggregated graph if at least one edge between $i$ and $j$ is present in one of the layers. Similarly, the minimum-based one corresponds to the case where edges are kept in the aggregated graph if they are present in all the layers. 
Clearly, these approaches are particularly effective when all the links in all the layers can be trusted, or when no layer can be trusted individually, respectively. However, in real-world applications, layers may contain complementary community information and some layers may be (partially) ``noisy'', i.e.\ they may contain limited or no information about the communities at all \cite{choobdar2019assessment,mercado2019generalized}.

\begin{table*}[t]
\caption{Entries of the generalized mean adjacency matrix $A(\boldsymbol{\theta})$, for particular choices of the parameters $(\alpha,\boldsymbol{\beta})\in \mathbb R^{K+1}$.}
\label{tab:gen_mean}
\centering
\begin{tabular}{cccccc}
\toprule
$\alpha \to -\infty $ & $\alpha = -1$, $\boldsymbol{\boldsymbol{\beta}}_k=1/K$ & $\alpha \to 0$, $\boldsymbol{\boldsymbol{\beta}}_k=1/K$ & $\alpha = 1$, $\boldsymbol{\boldsymbol{\beta}}_k=1/K$ & $\alpha \to +\infty$\\
 Minimum (MIN) & Harmonic (HARM) & Geometric (GEO) & Arithmetic (ARIT) &  Maximum (MAX) \\
\midrule
 $ \min_{k=1,\ldots,K} A_{ij}^{(k)}$ & 
 $  \left(\frac 1 K\sum_{k=1}^{K} \frac{1}{A_{ij}^{(k)}}\right)^{-1}$ & 
 $ \left(\prod_{k=1}^K A_{ij}^{(k)}\right)^{1/K}$ & 
 $  \frac 1 K\sum_{k=1}^{K} A_{ij}^{(k)}$ &
$ \max_{k=1,\ldots,K} A_{ij}^{(k)}$ \\[3pt]
\bottomrule
\end{tabular}
\vskip -1em
\end{table*}

The use of the parameters $\boldsymbol{\beta}_k$ in the linear model allows us to give different weights to the layers, if knowledge about their community information content is available. However, making an informed a-priori assessment on the presence of noise or on the different types of community structure across the layers can be very complicated \cite{choobdar2019assessment}. 
In order to overcome this modeling limitation, we design a nonlinear aggregation strategy that includes linear-, maximum-, and minimum-based aggregation strategies as special cases and learns optimal aggregation parameters from the available input labels. 


Both the linear combination $A^{lin}$ and the minimum, maximum matrices $A^{min}$, $A^{max}$ can be seen as  particular cases of more general nonlinear aggregations based on the  \textit{generalized mean adjacency matrix} 
$A(\alpha,\boldsymbol{\beta})$, entrywise defined~as 
\begin{equation}
\label{gen_mean}
    A(\alpha,\boldsymbol{\beta})_{ij}
    =  \left( \sum_{k=1}^{K} \boldsymbol{\beta}_k \big(A_{ij}^{(k)}\big)^\alpha \right)^{1/ \alpha}  ,
\end{equation}
where $\sum_k\boldsymbol{\beta}_k=1$, $\boldsymbol{\beta}_k>0$ as above, and  $\alpha \in \mathbb R$. In fact, $A^{lin}=A(1,\boldsymbol{\beta})$, and $A^{min}=\lim_{\alpha\to-\infty}A(\alpha,\boldsymbol{\beta})$,  $A^{max}=\lim_{\alpha\to+\infty}A(\alpha,\boldsymbol{\beta})$. As illustrated in Table \ref{tab:gen_mean}, the nonlinearity introduced by the parameter $\alpha$ allows us model a broad class of aggregation functions through \eqref{gen_mean}, including the maximum, the minimum, the harmonic, and the geometric means. Hence, using the generalized mean in \eqref{gen_mean} enables us to properly tune between all the different aggregation methods in Table \ref{tab:gen_mean}. This is particularly useful as linear aggregations alone may be inappropriate for a variety of multilayer network topologies. 
For example,  consider the case of a multilayer graphs with $K \gg 1$ layers, such that certain relevant edges appear only in one layer, which itself does not contain edges that are instead present in all the remaining layers. In this case, we need to consider the union (thus the maximum mean) of the edges across the layers, while their linear aggregation would most likely fail to capture the whole edge information. This is somehow shown by the ``complementary'' setting in our synthetic experiments in \S\ref{Synthetic Datasets}.

Note that, despite a similar terminology, $A(\alpha,\boldsymbol{\beta})$ is an elementwise  function and thus is very different from the matrix function generalized mean considered in \cite{mercado2019generalized}.



Letting $D(\alpha,\boldsymbol{\beta})=\mathrm{diag}(A(\alpha,\boldsymbol{\beta})\mathbf{1})$ be the  degree matrix of the generalized mean adjacency matrix, and $L(\alpha,\boldsymbol{\beta})=D(\alpha,\boldsymbol{\beta})-A(\alpha,\boldsymbol{\beta})$ its Laplacian matrix, we extend \eqref{eq:single_layer_multi_label_opt_problem} to the multilayer setting by considering the following
\[
\label{objfunmulti}
\textstyle{\varphi(X,Y;\alpha,\boldsymbol{\beta},\lambda) = \|X-Y\|_F^2 +\frac\lambda 2 \mathrm{Tr}(X^\top L(\alpha,\boldsymbol{\beta}) X)}
\]
and the corresponding class-wise function $\varphi_k(x,y;\alpha,\boldsymbol{\beta},\lambda)$, obtained by replacing $A$ with $A(\alpha,\boldsymbol{\beta})$ in \eqref{eq:single_layer_single_label_opt_problem}. 
In order to learn the parameters $\boldsymbol{\theta} \vcentcolon = (\alpha,\boldsymbol{\beta},\lambda$), we split the available input labels into training and test sets, with corresponding one-hot matrices $Y^{tr}$ and $Y^{te}$, and consider the bilevel optimization model
\begin{equation}
\label{bprob}
\begin{aligned}
\min_{\boldsymbol{\theta}} \quad & 
H(Y^{te},X_{Y^{tr};\boldsymbol{\theta}})\\
\textrm{s.t.} \quad &
X_{Y^{tr};\boldsymbol{\theta}} = 
\textstyle{\argmin_{X} \varphi(X,Y^{tr};\boldsymbol{\theta})}\\ 
\quad &
\boldsymbol{\theta}=(\alpha,\boldsymbol{\beta},\lambda), \; \alpha \in \mathbb{R}, \; 
\boldsymbol{\beta} \geq 0, \; \textstyle{\sum_k \boldsymbol{\beta}_k = 1},
\;\lambda \in \mathbb{R}
\end{aligned}
\end{equation}
where $H$ is the multiclass cross-entropy loss 
 \vspace{-.5em}
\begin{equation*}
    H(Y,X) = 
    -\frac{1}{N}
    \sum_{i=1}^{N}
    \sum_{j=1}^{N}
    Y_{ij}
    \log\left(\frac{X_{ij}}{\sum_{j=1}^{N} X_{ij}}\right).
\end{equation*}
The resulting embedding $X^*$ for the learned parameters is then used to classify the unlabeled nodes in the usual way. 

Note that, unlike the single-layer case, using \eqref{bprob_single}  rather than \eqref{bprob} in this setting may yield different results. In particular, if different layers carry information about different communities, using a one-vs-all cross-entropy model may be more effective. Thus, as an alternative to \eqref{bprob}, we consider 
\begin{equation}
\label{bprob_single}
\begin{aligned}
\min_{\boldsymbol{\theta}} \quad & 
h(y^{te},x_{y^{tr};\boldsymbol{\theta}})\\
\textrm{s.t.} \quad &
x_{y^{tr};\boldsymbol{\theta}} = \textstyle{
\argmin_{x} \varphi_k (x,y^{tr};\boldsymbol{\theta})}\\ 
\quad &
\boldsymbol{\theta}=(\alpha,\boldsymbol{\beta},\lambda), \; \alpha \in \mathbb{R}, \; 
\boldsymbol{\beta} \geq 0, \; \textstyle{\sum_k \boldsymbol{\beta}_k = 1},
\;\lambda \in \mathbb{R}
\end{aligned}
\end{equation}
which we solve for each community $k$, individually, using the binomial cross-entropy loss
\begin{equation*}
h(y,x) =
 -\frac{1}{N} \sum_{i=1}^{N} 
\big( y_i \log(x_i)
+ (1-y_i) \log(1 - x_i) \big) \, .
\end{equation*}

\section{Optimization with Inexact Gradient Computations}
In order to compute the generalized mean-based classifier, we use a gradient-free optimization algorithm combined with a form of parametric Label Propagation, as detailed in \S\ref{sub:impl}.
In particular, we consider a Frank-Wolfe (or conditional gradient) method~\cite{frank1956algorithm,jaggi2013revisiting,bomze2021frank} with inexact gradient and a tailored line search.
The method is described and analyzed below,
limiting our attention to~\eqref{bprob} for the sake of simplicity. Everything transfers straightforwardly to \eqref{bprob_single}. 

 Note that,
fixing the parameters $\boldsymbol{\theta} = (\alpha,\boldsymbol{\beta},\lambda)$, the inner problem $\min_X \varphi(X,Y;\boldsymbol{\theta})$ can be solved explicitly. In fact, a direct computation shows that 
$
\nabla_X \varphi(X,Y;\boldsymbol{\theta}) = 2\{(X-Y)+\lambda L(\alpha,\boldsymbol{\beta})\}X
$. 
Thus, 
\begin{equation}\label{eq:explicit_inner}
    \textstyle{\argmin_X \varphi(X,Y;\boldsymbol{\theta}) = \big(I+\lambda L(\alpha,\boldsymbol{\beta})\big)^{-1}Y\, .}
\end{equation}
Using \eqref{eq:explicit_inner}, we can  rewrite \eqref{bprob} by replacing the optimality constraint at the inner level with its explicit solution. Moreover, as the generalized mean converges fast to maximum and minimum for $\alpha\to\pm \infty$, we limit $\alpha$ within an interval $\alpha\in [-a,a]$, for a large enough $a>0$. Similarly, we restrict our study to $\lambda \in [l_0,l_1]$. 
Altogether, we reformulate \eqref{bprob} as 
\begin{equation}
\label{bprob2}
\begin{aligned}
 \min_{\boldsymbol{\theta}\in S} \; f(\boldsymbol{\theta})\, , 
\end{aligned}
\end{equation}
where $f(\boldsymbol{\theta})  := H(Y^{te},(I + {\lambda}L(\boldsymbol{\theta}))^{-1}Y^{tr})$ and, for $a>0$,  $S=\{(\alpha,\boldsymbol{\beta},\lambda)\in \mathbb R^{K+2}:\alpha\in[-a,a], \boldsymbol{\beta}_k>0,\sum_k\boldsymbol{\beta}_k=1,\lambda\in[l_0,l_1]\}$. 

As mentioned above, we use a Frank-Wolfe based method to solve~\eqref{bprob2}.
The rationale behind the algorithm is to compute, at every iteration $n$, a direction $d_n$ minimizing a linear approximation of $f$ around the current point $\boldsymbol{\theta}_n$. Then we obtain the next point $\boldsymbol{\theta}_{n+1}$ by moving along $d_n$ with a stepsize $\eta_n$, chosen by a proper line search.
Although $f$ is a smooth real-valued function, the computation of its gradient $\nabla f$ can be extremely expensive in practice. To overcome this issue, instead of $\nabla f$, in the algorithm we use an estimate $\widetilde \nabla f$.
The resulting method is presented in Algorithm~\ref{alg:final_FW}. 

\begin{algorithm}[!t]
\caption{Frank-Wolfe algorithm with inexact gradient}
\label{alg:final_FW}
\begin{algorithmic}[1]
\STATE \textbf{Given} $\theta_0 \in S$ 
\STATE \textbf{For} $n = 0,1,\ldots$ 
\STATE \hspace*{0.5truecm}Compute $\widetilde \nabla f(\theta_n)$ as an estimate of $\nabla f(\theta_n)$ \label{linear_prob}
\STATE \hspace*{0.5truecm}Compute $\hat \theta_n \in \argmin_{\theta \in S}\widetilde \nabla f(\theta_n)^\top  (\theta-\theta_n)$\\\hspace*{0.5truecm} and set $d_n = \hat \theta_n - \theta_n$ \label{line:linearized_FW}
\STATE \hspace*{0.5truecm}Compute a stepsize $\eta_n\in(0,1]$ by a line search
\STATE \hspace*{0.5truecm}Set $\theta_{n+1} = \theta_n + \eta_n d_n$ 
\STATE \textbf{End for}
\end{algorithmic}
\end{algorithm}

%
%

Note that the linear problem at line~\ref{linear_prob} of Algorithm~\ref{alg:final_FW} is particularly simple due to the box-plus-simplex form of the constraint set $S$, and it can be solved separately in  the variables $\alpha$, $\boldsymbol{\beta}$, and $\lambda$. 
In fact, for the variable $\alpha$, we aim at minimizing a linear function over the box $[-a,a]$, which implies $\hat{\alpha}_n =-a$ if $\tilde \nabla_{\alpha}f(\boldsymbol{\theta}_n) > 0$, and $\hat{\alpha}_n =a$ otherwise.
Similarly, for the variable $\lambda$, we set  $\hat{\lambda}_n =l_0$ if $\tilde \nabla_{\lambda}f(\boldsymbol{\theta}_n) > 0$, and $\hat{\lambda}_n =l_1$ otherwise.
For the variables $\boldsymbol{\beta} \in \mathbb{R}^K$, we have to minimize a linear function over the unit simplex, which yields $\boldsymbol{\hat\beta}_n = 
e_{\hat{\jmath}}$, 
where
$\hat{\jmath} = \argmin_{j=1,\ldots,K}
[(\tilde \nabla_{\boldsymbol{\beta}}f(\boldsymbol{\theta}_n)]_j$
and $e_{\hat{\jmath}}$ is the $\hat{\jmath}$-th vector of the canonical basis of $\mathbb{R}^K$.

\subsection{Convergence Analysis}




To analyze the convergence of Algorithm~\ref{alg:final_FW}, we first introduce some useful notation.
Let $g_n = -\nabla f(\boldsymbol{\theta}_n)^\top d_n$, $\tilde g_n = -\widetilde\nabla f(\boldsymbol{\theta}_n)^\top d_n$ and $g_n^{FW} = -\nabla f(\boldsymbol{\theta}_n)^\top d_n^{FW}$, 
where $d_n^{FW} \in \argmin_{\boldsymbol{\theta} \in S}\{\nabla f(\boldsymbol{\theta}_n)^\top (\boldsymbol{\theta}-\boldsymbol{\theta}_n)\} - \boldsymbol{\theta}_n$ is the direction obtained by the Frank-Wolfe algorithm with exact gradient.
Inspired by~\cite{freund2016new}, we assume that the estimate $\widetilde \nabla f$ satisfies the following condition.

\begin{assumption}\label{assump:cond_inexact}
For every $n$, there exists $\epsilon_n\geq 0$ such that 
\begin{equation}\label{cond_inexact}
|(\nabla f(\boldsymbol{\theta}_n) - \widetilde \nabla f(\boldsymbol{\theta}_n))^\top (\boldsymbol{\theta}-\boldsymbol{\theta}_n)| \leq \epsilon_n \quad \forall\  \boldsymbol{\theta} \in S.
\end{equation}
\end{assumption}
Since $S$ is a convex set, a point $\boldsymbol{\theta^*} \in S$ is said to be stationary for~\eqref{bprob2} when $\nabla f(\boldsymbol{\theta^*})^\top (\boldsymbol{\theta}-\boldsymbol{\theta^*}) \ge 0$ for all $\boldsymbol{\theta} \in S$. Then, $g^{FW}_n$ is an optimality measure, i.e.\ $g^{FW}_n=0$ if and only if $\boldsymbol{\theta}_n\in S$ is a stationary point. 
Now we show that, when Assumption~\ref{assump:cond_inexact} is satisfied
with a sufficiently small $\epsilon_n$ and the stepsize $\eta_n$ is generated with a suitable line search, Algorithm~\ref{alg:final_FW} 
obtains a stationary point at a sublinear rate on non-convex objectives with a Lipschitz continuous gradient.  The constant in the convergence rate   depends on the quality of the gradient estimate (the more precise the estimate, the smaller the constant). The proof can be found in \cref{AppendixA}.
\begin{theorem} \label{nonconvb}
Let $\nabla f$ be Lipschitz continuous with constant $M$, and let $S$ be compact with finite diameter $\Delta$. Let $\{\boldsymbol{\theta}_n\}$ be a sequence generated by Algorithm~\ref{alg:final_FW}, where $\widetilde \nabla f$ satisfies Assumption~\ref{assump:cond_inexact} with \begin{equation}
 \label{assumption}
{\epsilon_n \leq \frac{\sigma}{1+\sigma}\,  \tilde g_n, \quad 0 \le \sigma < \frac{1}{3},}
 \end{equation} 
and  the step size $\eta_n$ satisfies
 \begin{equation}    \label{alphabound}
	{{\eta}_n \ge \bar{\eta}_n = \min\left(1, \frac{\tilde{g}_n}{{M\n{d_n}^2}} \right),}
\end{equation}
	\begin{equation} \label{eq:rho}
	f(\boldsymbol{\theta}_n) - f(\boldsymbol{\theta}_n + \eta_n d_n) \geq \rho\bar{\eta}_n\tilde{g}_n,
	\end{equation}
	with some fixed $\rho > 0$. 
Then, 
	\begin{equation} \label{g_T}
		{g_ n^* \leq \max\left(\sqrt{\frac{\Delta^2 M (f(\boldsymbol{\theta}_0) - f^*)}{n \rho (1 - \sigma)^2}}, \frac{2(f(\boldsymbol{\theta}_0) - f^*)}{n(1 - 3\sigma)} \right),}
	\end{equation}
 where $\displaystyle g^*_n = \min_{0 \leq i \leq n-1} g^{FW}_i$ 
 and $f^* = \min_{\boldsymbol{\theta} \in S} f(\boldsymbol{\theta})$.
\end{theorem}
Note that, in our setting, $\Delta \leq 2a+\sqrt 2$. Condition~\eqref{assumption} can be easily satisfied by a proper calculation of the gradient estimate $\widetilde \nabla f$ (see \S\ref{sub:impl}). 
Conditions~\eqref{alphabound}--\eqref{eq:rho} can  be  satisfied with suitable  line searches/stepsize rules (see, e.g., \cite{bomze2020active,bomze2021frank,rinaldi2022avoiding}). In particular, Lemma~\ref{lemma} below shows that this is the case for the modified Armijo line search rule which sets
\begin{equation}\label{eta}
\eta_n = \delta^j,
\end{equation}
where $j$ is the smallest non-negative integer such that
\begin{equation}
\label{Armijo}
f(\boldsymbol{\theta}_n) - f(\boldsymbol{\theta}_n + \eta_n d_n) \geq \gamma \eta_n \tilde{g}_n,
\end{equation}
with $\gamma\in (0,1/2)$ and $\delta \in (0,1)$ being two fixed parameters. The proof can be found in \cref{AppendixB}.
%
%
\begin{lemma} 
\label{lemma}
Let Assumption
\ref{assump:cond_inexact} hold with
\begin{equation}
\label{assumption2}
{\epsilon_n \leq \frac{\sigma}{1+\sigma} \, \tilde g_n, \quad 0 \le \sigma < \frac{1}{2}.}
\end{equation}
At iteration $n$, if $\eta_n$ is determined by the Armijo line search described in~\eqref{eta}--\eqref{Armijo}, then 
	\begin{equation} \label{Arbaralpha2}
		\eta_n \ge 
  \min\{1, 2\delta(1 - \gamma - \sigma)\} \bar{\eta}_n,
	\end{equation}	
	with $\bar \eta_n$ being defined as in~\eqref{alphabound}.
\end{lemma} 

Note that in the proof of~\cref{lemma} we prove
 \begin{equation*}    
	{
 {\eta}_n \geq  \min\left(1, c \frac{\tilde{g}_n}{{M\n{d_n}^2}} \right) \text{for some } c>0
 }
\end{equation*}
for the Armijo line search. When $c \geq 1$ then $\bar{\eta}_n$ is of course a lower bound for the step size $\eta_n$, and when $c < 1$ we can still recover \eqref{alphabound} by considering $\tilde{M} = M/c$ instead of $M$ as Lipschitz constant. 
The complexity analysis of the method can be obtained straightforwardly from \cref{nonconvb}. Details are reported in \cref{AppendixD}.

\subsection{Implementation Details}\label{sub:impl}
In Algorithm \ref{alg:final_FW}, we approximate the gradient with the finite difference method:
\begin{equation}\label{fd}
    \widetilde\nabla f(\boldsymbol{\theta}_n) = \sum_{i=1}^{K+2} \frac{f(\boldsymbol{\theta}_n + h_n e_i) - f(\boldsymbol{\theta}_n)}{h_n} e_i,
\end{equation}
where $h_n$ is a suitably chosen positive parameter. As shown in \cite{berahas2022theoretical}, this approach gives good gradient approximations in practice. 

 From \cite{sahu2019towards}, we have that \cref{cond_inexact} is satisfied when using the finite difference approach with $\epsilon_n = \frac{M\Delta (2+K)}{2} h_n$. We notice that condition \eqref{assumption} can in turn be satisfied at each iteration $k$ by suitably choosing  $h_n$ in the finite difference approximation, e.g.,   $h_n\leq \xi \tau$, with $\xi=\frac{2\sigma}{(1+\sigma) M\Delta (2+K)}$ and $\tau$  stopping condition tolerance.

 In our experiments, we start with $h_0 = 10^{-4}$ and set $h_{n} = \frac{h_{n-1}}{2}$ for $n = 1,2,\ldots$. Furthermore, we stop the algorithm when $\tilde g_n= -\widetilde \nabla f(\boldsymbol{\theta}_n)^\top  d_n \le \tau$, with $\tau=10^{-4}$.

In order to compute $f(\boldsymbol{\theta})$ for a given set of parameters $\boldsymbol{\theta}=(\alpha,\boldsymbol{\beta},\lambda)$, we run a form of modified parametric Label Propagation algorithm:
\begin{multline*}
X^{(r+1)} = {\lambda}A(\alpha,\boldsymbol{\beta})(I + {\lambda}D(\alpha,\boldsymbol{\beta}))^{-1}X^{(r)} +\\
+ (I + {\lambda}D(\alpha,\boldsymbol{\beta}))^{-1}Y \, ,
\end{multline*}
 which propagates the input labels in $Y$ and converges to the solution of the linear system \eqref{eq:explicit_inner}. 
In fact, as $\sum_{j}A(\alpha,\boldsymbol{\beta})_{ij}=D(\alpha,\boldsymbol{\beta})_{ii}$ for all $i$, a direct application of the first Gershgorin circle theorem \cite{varga2010gervsgorin} to the matrix $A(\alpha,\boldsymbol{\beta})(I + {\lambda}D(\alpha,\boldsymbol{\beta}))^{-1}$  implies that the spectral radius of $A(\alpha,\boldsymbol{\beta})(I + {\lambda}D(\alpha,\boldsymbol{\beta}))^{-1}$ is smaller than one and thus $X^{(r)}\to \argmin_X\varphi(X,Y;\boldsymbol{\theta})$, as $r\to\infty$.

We apply the multistart version of Frank-Wolfe \cite{marti2013multi}, where the algorithm is applied with different initial points, and we choose the best solution according to the value of the optimized function $f$. In particular, we start from 10 random points $\boldsymbol{\theta}_0$, among which we include the particular choices $\boldsymbol{\theta}_0=(1,1/K,\dots,1/K,1)$ and $\boldsymbol{\theta}_0=(-1,1/K,\dots,1/K,1)$, which correspond to the arithmetic and the harmonic means. 
In all the experiments, we restricted the study of the parameters  $\alpha$ and $\lambda$ as follows: $\alpha \in [-20,20]$ and $\lambda \in [0.1,10]$. The latter boils down to the standard $[0.1,0.9]$ search interval for the variable $\lambda/(1+\lambda)$, usually employed in label propagation algorithms. For the methods obtained with a fixed choice of  generalized mean aggregation function (see Table \ref{tab:gen_mean} for  details), we fixed $\lambda = 1$.


\section{Numerical Results}
\label{Experiments}
We perform tests on different synthetic and real-world multilayer networks. For each data set, we are given an input set of known labels $Y$, which we split into a train $Y^{tr}$ set with 80\% of the available labels, and a test  $Y^{te}$ set, formed by the remaining 20\%. Since the resulting optimal parameters can change depending on the selection of the training and test sets, we initially randomly split the input labeled nodes into 5 sets of equal size and then cyclically assign one of the sets to $Y^{te}$ and the remaining ones to $Y^{tr}$. For each of these choices, we run Algorithm \ref{alg:final_FW} with 10 starting points as discussed in \S\ref{sub:impl}, and we eventually choose the parameters that yield the lowest value of the loss function $f(\boldsymbol{\theta})$ over the test set among the five runs. 
Once the optimal weights are computed, we run standard Label Propagation on the resulting aggregated graph and we asses the accuracy performance  on the held-out test set, which is comprised of all the initially
non-labeled points.


We implement both the multiclass (MULTI) and the binomial (BINOM) versions of our method, corresponding to the bilevel optimization problems in \eqref{bprob}  and \eqref{bprob_single}, respectively. Our \texttt{Python} implementation is available at
\small
\url{https://github.com/saraventurini/Learning-the-right-layers-on-multilayer-graphs}.
\normalsize

We considered both synthetic and real-world networks, performing extensive experiments to compare the proposed approach,  which learns the parameters of the generalized mean from the available input data, against standard Label propagation on each single layer, as well as the methods corresponding to the proposed generalized mean aggregation function for some special choices of the parameters (those illustrated in Table \ref{tab:gen_mean}), and  four multilayer graph semi-supervised learning baselines:

\begin{itemize}[noitemsep, topsep=0pt,leftmargin=*]
\item \textbf{SGMI}: Sparse Multiple Graph Integration \cite{karasuyama2013multiple}, based on label propagation by sparse integration, with parameters $\lambda_1 = 1, \lambda_2 = 10^{-3}$;
\item \textbf{AGML}: Auto-weighted Multiple Graph
Learning \cite{nie2016parameter}, which is a parameter-free method for optimal graph
layer weights;
\item \textbf{SMACD}: Semi-supervised Multi-Aspect Community Detection~\cite{gujral2018smacd}, which
is a tensor factorization method for semi-supervised learning;
\item \textbf{GMM}: Generalized Matrix Means, which is a Laplacian-regularization approach based on the Log-Euclidean matrix function formulation of the power mean Laplacian, with parameter $p=-1$~\cite{mercado2019generalized};
\end{itemize}
Notice that SGMI and GMM need a parameter choice, which we have made following the indications in the corresponding papers, while AGML and SMACD, as well as the proposed methods MULTI and BINOM,  are parameter-free. 
We also tested against the two multilayer graph neural networks discussed in \S\ref{Related Work}, which however performed poorly, probably due to the absence of features in our test settings.


\begin{figure}[t]
\label{fig:synt}
\centering
           \subfigure{%
              \includegraphics[width=\columnwidth, trim={3cm 0.5cm 3cm 1cm},clip]{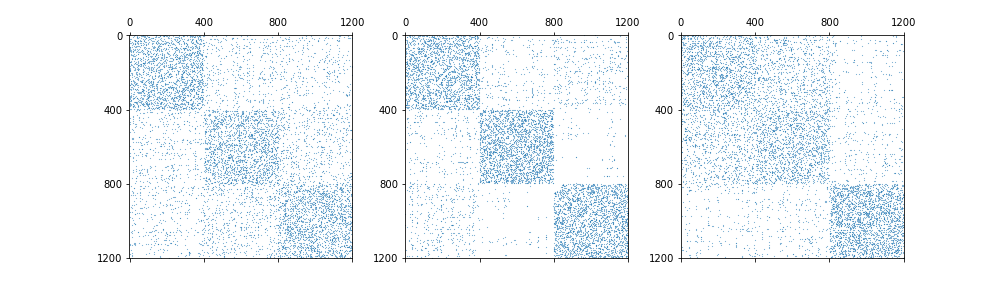}%
              \label{fig:info}%
           } 
           \vskip -.5em
           \subfigure{%
              \includegraphics[width=\columnwidth, trim={3cm 0.5cm 3cm 1cm},clip]{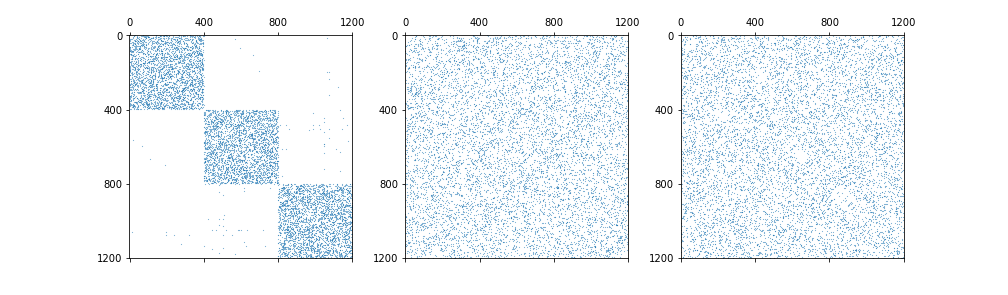}%
              \label{fig:noisy}%
           }
           \vskip -.5em
           \subfigure{%
              \includegraphics[width=\columnwidth, trim={3cm 0.5cm 3cm 1cm},clip]{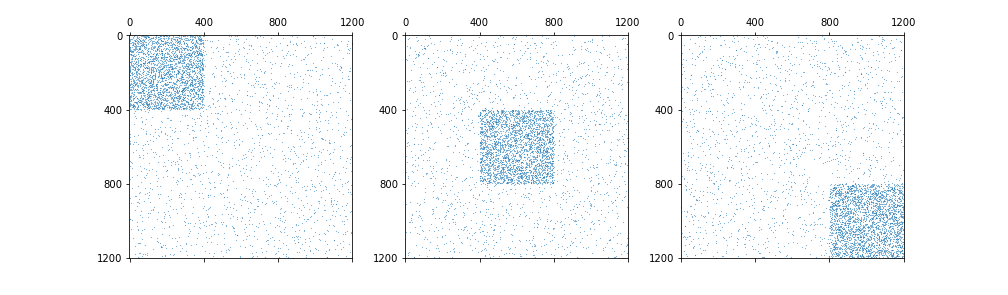}%
              \label{fig:compl}%
           }
           \vskip -.5em
           \caption{Synthetic datasets settings. (top) informative case, (middle) noisy case, (bottom) complementary case.}
           \label{fig:default}
\vskip -2em
\end{figure}  

\subsection{Synthetic Datasets}
\label{Synthetic Datasets}
We created synthetic datasets with 3 communities of 400 nodes each, and 3 layers.
In particular, for each layer, we generated 3 isotropic Gaussian blobs of points $p_i\in \mathbb R^5$, 
with a variable standard deviation. 
The adjacency matrix of the network is then formed by means of a  symmetrized  $k$-NN graph with $k = 5$, weighted with the Euclidean kernel $\exp(-\|p_i-p_j\|+\min_{ij}\|p_i-p_j\|)$. 
We considered three settings (illustrated in Figure~\ref{fig:synt}):
\begin{itemize}[noitemsep, topsep=0pt,leftmargin=*]
    \item \textit{Informative case}: layers are formed by 3 isotropic Gaussian blobs and all show the same community~structure; 
    \item \textit{Noisy case}: one layer is informative and the other two are noise. The noisy layers are generated by a random reshuffling of the informative ones. 
    \item \textit{Complementary case}: each layer carries  information concerning only one cluster, while is noisy for the remaining ones. The noise layers are sparser here than in the previous setting (we shuffle $k$-NN layers with $k=1$).   
\end{itemize}
The informative isotropic Gaussian blobs have standard deviation $std \in \{5,6,7,8\}$ for the informative case, as this is the easiest setting, while we test for $std \in \{2,3,4,5\}$ in the other two settings, as these are more challenging. 
The percentage of input labels is $20\%$ of the overall number of nodes in each of the communities. 

Table \ref{tab:info2} reports the average accuracy and standard deviation score across 5 network samples, as compared to the accuracy of the individual layers (computed ignoring the other layers), reported in the first three columns, and those achieved with fixed a-priori choices of the parameters (as in Table~\ref{tab:gen_mean}).
%
%
%
%
%
%
%
%
%
The proposed BINOM and MULTI perform well across all settings, most of the time outperforming  each individual layer and the considered baselines. 
In particular, in all the settings, MIN, GEO and SMACD perform  poorly. AGML works well mostly in the informative setting. ARIT, HARM, and MAX show good performances in the informative and complementary cases, but not in the noisy one. SGMI achieves high accuracy only in the noisy case. 
GMM performs well only in the informative and noisy settings. 

While the best performance is sometimes achieved by  some particular aggregation function (such as MAX or HARM), all the baselines are setting-specific and have poor performance in certain settings, e.g. in the presence of noise. When measured across all settings,  BINOM and MULTI perform best. This is  highlighted by the Average Performance Ratio (\textrm{APR}) score values, reported in the last line of Table~\ref{tab:info2}, which are quantified as follows:  denoting the accuracy of algorithm $a$ on dataset $d$ as $\mathcal A_{a,d}$,
 let the {performance ratio} be  $r_{a,d}=\mathcal A_{a,d}/\max\{\mathcal A_{a,d}\text{ over all }   a\}$. The APR of each algorithm  is then obtained by averaging $r_{a,d}$ over all the datasets $d$.
For any algorithm, the closer the average performance ratio is to 1, the better the overall performance.
We further compare the performances using the average rank (\textrm{AR}), computed by assigning $1$ to the best-performing method, $2$ to the second best, and so on, for each dataset. We  highlight  that the \textrm{APR} is a more informative metric as it also takes into account the value of the accuracy of the individual tests, while \textrm{AR} does not.

\setlength{\tabcolsep}{2.5pt}
\begin{table}[t]
\caption{Example of learned parameters by BINOM (B) and MULTI (M) on the synthetic datasets of Table~\ref{tab:info2}.}
\label{tab:learned_parameters}
\centering
\begin{scriptsize}
\resizebox{\columnwidth}{!}{
\begin{NiceTabular}{cc ccccc  ccccc  ccccc}
\toprule
 & & \multicolumn{5}{c}{INFO} & \multicolumn{5}{c}{NOISY} & \multicolumn{5}{c}{COMPL}\\
\cmidrule(lr){3-7} \cmidrule(lr){8-12} \cmidrule(lr){13-17}
 & $k$ & $\boldsymbol{\beta}_1$ & $\boldsymbol{\beta}_2$ & $\boldsymbol{\beta}_3$ & $\alpha$ & $\lambda$
 & $\boldsymbol{\beta}_1$ & $\boldsymbol{\beta}_2$ & $\boldsymbol{\beta}_3$ & $\alpha$ & $\lambda$
 & $\boldsymbol{\beta}_1$ & $\boldsymbol{\beta}_2$ & $\boldsymbol{\beta}_3$ & $\alpha$ & $\lambda$\\
\midrule
 B & 1\; & 0.33 & 0.32 & 0.35 & -2.8 & 1.12 & 1.0 & 0.0 & 0.0 & 20.0 & 10.0 & 0.83 & 0.12 & 0.05 & 10.27 & 8.62\\  
   & 2\; & 0.32 & 0.46 & 0.22 & -1.89 & 6.36 &  1.0  & 0.0 & 0.0 & 20.0 & 10.0 & 0.13 & 0.50 & 0.37 & 10.35 & 3.62\\    
   & 3\; & 0.23 & 0.36 & 0.41 & -3.54 & 1.02 & 1.0 & 0.0 & 0.0 & 20.0 & 10.0 & 0.11 & 0.24 & 0.65 & 10.95 & 4.69\\    
 M & -\; & 0.31 & 0.34 & 0.35 & -3.65 & 1.74 &  1.0 & 0.0 & 0.0 & 5.03 & 0.97 & 0.30 & 0.39 & 0.31 & -6.96 & 0.67\\ 
\bottomrule
\end{NiceTabular}
}
\end{scriptsize}
\vskip-1em
\end{table}

Moreover, by inspecting the learned weights, the proposed methods allow us to make an  assessment of the structure of the multilayer network and the presence of noisy or less informative layers. This is illustrated in Table~\ref{tab:learned_parameters}, where we show  an example of learned weights resulting from Algorithm \ref{alg:final_FW} 
Notice that, looking at the $\boldsymbol{\beta}$ parameter: in the informative case, the weight is equally distributed among the layers; in the noisy case, the methods consider just the informative layer disregarding completely the noisy layers; in the complementary case, MULTI distributes the weight equally among the layers, while BINOM gives a higher weight to the community correspondent to each layer. 

In \cref{AppendixD} we report a CPU-time  comparison with the methods in Table~\ref{tab:info2} as the number of nodes increases, which highlights the efficiency of our framework.

\subsection{Real World Datasets}
\label{Real World Datasets}
We consider nine real-world datasets frequently used to assess  performance of multilayer graph clustering ~\cite{mercado2019generalized,magnani2021community,venturini2022variance}:
\begin{itemize}[noitemsep, topsep=0pt,leftmargin=*]
\item \textit{3sources}:  news
articles  covered by news sources BBC, Reuters, and Guardian (169 nodes, 6 communities, 3 layers)~\cite{greene2009matrix,liu2013multi};
\item \textit{BBC}: BBC news articles (685 nodes, 5 communities, 4 layers)~\cite{greene2005producing};
\item \textit{BBCSport}:  BBC Sport articles (544 nodes, 5 communities, 2 layers)~\cite{greene2009matrix};
\item \textit{Wikipedia}:  Wikipedia articles (693 nodes, 10 communities, 2 layers)~\cite{rasiwasia2010new};
\item \textit{UCI}:  hand-written UCI digits dataset with six different sets of features (2000 nodes, 10 communities, 6 layers)~\cite{dua2017uci,liu2013multi};
\item \textit{cora}:  citations dataset (2708 nodes, 7 communities, 2 layers)~\cite{mccallum2000automating};
\item \textit{citeseer}:  citations dataset (3312 nodes, 6 communities, 2 layers)~\cite{lu2003link};
\item \textit{dkpol}:  five types of relationships between employees of a university department (490  nodes, 10 communities, 3 layers)~\cite{hanteer2018interaction};
\item \textit{aucs}:  three types of online relations between
Danish Members of the Parliament on Twitter (61 nodes, 9 communities, 5 layers)~\cite{rossi2015towards}.
   \end{itemize}
For each dataset, we assume either $1\%$ or $10\%$ of the labels are initially known, for each class. 
Tables~\ref{tab:1layer1percent} and \ref{tab:1layer10percent} report the average accuracy ($\pm$ standard deviation) across 3 samples of the known labels. 
 In addition to the original datasets, we show the performance when one additional noisy layer is added.   In Appendix~\ref{sec:additional_results} we report additional results for different known label percentages and with the addition of  two layers of noise. 
We do not compare the methods against the AGML baseline as that method is designed  for graphs with communities of the same size. 
The results confirm the same behavior observed in the synthetic case, showing that  BINOM and MULTI match or overcome the baselines in most cases, and are the best-performing methods overall, across all settings. 
We also emphasize that BINOM and MULTI are the only techniques that consistently match or outperform the single layers used in isolation, showing that our approach is able to effectively take advantage of the multilayer structure in all different settings. This is a particularly remarkable and desirable property, which is directly related to a recent data challenge \cite{choobdar2019assessment}.

In \cref{AppendixF} we report a table with the different parameters learned by the methods. The numbers are averaged over three random samplings of the initially labeled nodes.

\vspace{-0.8em}

\section{Conclusion}
\label{Conclusion and Future Works}
We proposed a parameter-free method for semi-supervised classification on multiplex networks that identifies relevant layers by learning a nonlinear aggregation function from  the known labels. 
We formulate the model as a bilevel optimization problem which we solve  using an inexact Frank-Wolfe algorithm combined with a parametric Label Propagation scheme. We provide a detailed convergence analysis of the method. 
Experimental results against single-layer approaches and a variety of baselines,  on both synthetic and real-world datasets, demonstrate that the proposed method is able to identify relevant layers and thus obtain consistent 
 and robust performance across different clustering settings,  in particular when some layers are mostly just noise. 

\def\inrule{ \specialrule{.01pt}{1pt}{1pt} }
\setlength{\tabcolsep}{2.5pt}
\begin{table*}[t]
\caption{Accuracy (mean $\pm$ standard deviation) over five random samples of synthetically generated multilayer graphs, for different levels of $std$ in the isotropic Gaussian blobs forming the clusters.}
\label{tab:info2}
\centering
\begin{scriptsize}
\resizebox{\textwidth}{!}{
\begin{NiceTabular}{cc | ccc | ccccc | cc | cccc}
\toprule
 & $std$ & \RNum{1} & \RNum{2} & \RNum{3} & MIN & GEOM & ARIT & HARM & MAX & BINOM & MULTI & SGMI & AGML & SMACD & GMM\\
\midrule
\multirow{4}{*}{\rotatebox[origin=c]{90}{INFO}} &        5 & 0.83$\pm$0.1 & 0.86$\pm$0.05 & 0.84$\pm$0.1 & 0.33$\pm$0.0 & 0.33$\pm$0.0 & 0.95$\pm$0.03 & 0.96$\pm$0.03 & 0.96$\pm$0.03 & 0.97$\pm$0.02 & 0.93$\pm$0.04 & 0.87$\pm$0.04 & 0.93$\pm$0.05 & 0.48$\pm$0.18 & 0.93$\pm$0.04\\
&        6 & 0.77$\pm$0.11 & 0.8$\pm$0.05 & 0.79$\pm$0.11 & 0.33$\pm$0.0 & 0.33$\pm$0.0 & 0.9$\pm$0.05 & 0.92$\pm$0.04 & 0.92$\pm$0.05 & 0.91$\pm$0.06 & 0.87$\pm$0.07 & 0.8$\pm$0.06 & 0.86$\pm$0.09 & 0.47$\pm$0.16 & 0.89$\pm$0.05\\ 
&        7 & 0.71$\pm$0.1 & 0.74$\pm$0.05 & 0.74$\pm$0.11 & 0.33$\pm$0.0 & 0.33$\pm$0.0 & 0.86$\pm$0.07 & 0.88$\pm$0.06 & 0.87$\pm$0.07 & 0.87$\pm$0.06 & 0.83$\pm$0.08 & 0.74$\pm$0.07 & 0.8$\pm$0.1 & 0.46$\pm$0.14 & 0.85$\pm$0.07\\
&        8 & 0.66$\pm$0.1 & 0.69$\pm$0.04 & 0.7$\pm$0.11 & 0.33$\pm$0.0 & 0.33$\pm$0.0 & 0.81$\pm$0.08 & 0.83$\pm$0.08 & 0.82$\pm$0.09 & 0.8$\pm$0.09 & 0.78$\pm$0.08 & 0.69$\pm$0.07 & 0.74$\pm$0.1 & 0.44$\pm$0.14 & 0.8$\pm$0.08 \\ 
\midrule
\multirow{4}{*}{\rotatebox[origin=c]{90}{NOISY}} &        2 & 0.99$\pm$0.02 & 0.35$\pm$0.01 & 0.36$\pm$0.01 & 0.33$\pm$0.0 & 0.33$\pm$0.0 & 0.67$\pm$0.02 & 0.68$\pm$0.03 & 0.68$\pm$0.03 & 0.99$\pm$0.01 & 0.99$\pm$0.02 & 0.98$\pm$0.03 & 0.63$\pm$0.03 & 0.35$\pm$0.03 & 0.86$\pm$0.04 \\ 
&        3 & 0.95$\pm$0.05 & 0.35$\pm$0.0 & 0.36$\pm$0.01 & 0.33$\pm$0.0 & 0.33$\pm$0.0 & 0.66$\pm$0.04 & 0.66$\pm$0.04 & 0.66$\pm$0.04 & 0.96$\pm$0.04 & 0.95$\pm$0.05 & 0.94$\pm$0.07 & 0.6$\pm$0.06 & 0.36$\pm$0.04 & 0.82$\pm$0.07 \\ 
&        4 & 0.89$\pm$0.08 & 0.36$\pm$0.0 & 0.35$\pm$0.01 & 0.33$\pm$0.0 & 0.33$\pm$0.0 & 0.63$\pm$0.05 & 0.63$\pm$0.05 & 0.63$\pm$0.05 & 0.9$\pm$0.08 & 0.89$\pm$0.08 & 0.88$\pm$0.1 & 0.57$\pm$0.07 & 0.36$\pm$0.03 & 0.77$\pm$0.1 \\
&        5 & 0.83$\pm$0.1 & 0.35$\pm$0.01 & 0.36$\pm$0.01 & 0.33$\pm$0.0 & 0.33$\pm$0.0 & 0.58$\pm$0.06 & 0.59$\pm$0.06 & 0.58$\pm$0.06 & 0.85$\pm$0.09 & 0.83$\pm$0.1 & 0.8$\pm$0.12 & 0.53$\pm$0.07 & 0.35$\pm$0.04 & 0.7$\pm$0.11 \\ 
\midrule
\multirow{4}{*}{\rotatebox[origin=c]{90}{COMPL}} &        2 & 0.41$\pm$0.0 & 0.47$\pm$0.0 & 0.48$\pm$0.01 & 0.33$\pm$0.0 & 0.33$\pm$0.0 & 0.89$\pm$0.02 & 0.94$\pm$0.02 & 0.92$\pm$0.02 & 0.86$\pm$0.01 & 0.89$\pm$0.02 & 0.43$\pm$0.04 & 0.3$\pm$0.01 & 0.52$\pm$0.05 & 0.69$\pm$0.03 \\
&        3 & 0.41$\pm$0.01 & 0.47$\pm$0.01 & 0.47$\pm$0.01 & 0.33$\pm$0.0 & 0.33$\pm$0.0 & 0.86$\pm$0.04 & 0.9$\pm$0.05 & 0.89$\pm$0.05 & 0.85$\pm$0.02 & 0.87$\pm$0.04 & 0.44$\pm$0.04 & 0.3$\pm$0.02 & 0.44$\pm$0.08 & 0.67$\pm$0.04 \\ 
&        4 & 0.4$\pm$0.01 & 0.46$\pm$0.01 & 0.47$\pm$0.01 & 0.33$\pm$0.0 & 0.33$\pm$0.0 & 0.82$\pm$0.06 & 0.86$\pm$0.06 & 0.85$\pm$0.06 & 0.82$\pm$0.01 & 0.83$\pm$0.06 & 0.43$\pm$0.04 & 0.18$\pm$0.16 & 0.71$\pm$0.17 & 0.65$\pm$0.05 \\ 
&        5 & 0.4$\pm$0.01 & 0.46$\pm$0.02 & 0.47$\pm$0.01 & 0.33$\pm$0.0 & 0.33$\pm$0.0 & 0.79$\pm$0.07 & 0.82$\pm$0.08 & 0.81$\pm$0.07 & 0.81$\pm$0.07 & 0.8$\pm$0.07 & 0.43$\pm$0.04 & 0.18$\pm$0.17 & 0.46$\pm$0.13 & 0.63$\pm$0.06 \\ 
\midrule
\multicolumn{2}{c}{\textrm{APR}}   & - &-&- & 0.37& 0.37 & 0.87 & 0.90 & 0.89 & \textbf{0.98} & 0.97 & 0.78 & 0.61 & 0.50 &  0.86 \\
\multicolumn{2}{c}{\textrm{AR}}   & - &-&- & 10.0& 10.0 & 5.0 & 2.9 & 3.5 & \textbf{2.5} & 3.8 & 6.1 & 8.6 & 8.7 & 5.1 \\
 \bottomrule
\end{NiceTabular}
}
\end{scriptsize}
\vskip -1.5em
\end{table*}

\begin{table*}[t]
\caption{Accuracy (mean $\pm$ std) over three random samples of the 1\% of input labels, on real-world datasets (+ one layer of noise).}
\label{tab:1layer1percent}
\centering
\begin{scriptsize}
\resizebox{\textwidth}{!}{
\begin{NiceTabular}{c|cccccc|cccccc|cc|ccc}
\toprule
 & \RNum{1} & \RNum{2} & \RNum{3} & \RNum{4} & \RNum{5} & \RNum{6} &  & MIN & GEOM & ARIT & HARM & MAX & BINOM & MULTI & SGMI & SMACD & GMM\\
\midrule
\multirow{2}{*}{3sources}&\multirow{2}{*}{0.72}&\multirow{2}{*}{0.69}&\multirow{2}{*}{0.77}&\multirow{2}{*}{-}&\multirow{2}{*}{-}&\multirow{2}{*}{-}& & 0.66$\pm$0.01&0.66$\pm$0.01&0.79$\pm$0.04&0.71$\pm$0.04&0.76$\pm$0.04&0.79$\pm$0.04&0.79$\pm$0.04&0.65$\pm$0.04&0.58$\pm$0.06&0.69$\pm$0.06 \\
 & & & & & & & (+1 noise) &0.35$\pm$0.01&0.35$\pm$0.01&0.73$\pm$0.06&0.55$\pm$0.05&0.69$\pm$0.07&0.73$\pm$0.06&0.73$\pm$0.06&0.52$\pm$0.22&0.65$\pm$0.08&0.62$\pm$0.02\\
 \inrule
\multirow{2}{*}{BBC}&\multirow{2}{*}{0.77}&\multirow{2}{*}{0.76}&\multirow{2}{*}{0.72}&\multirow{2}{*}{0.75}&\multirow{2}{*}{-}&\multirow{2}{*}{-}& & 0.35$\pm$0.01&0.35$\pm$0.01&0.83$\pm$0.01&0.81$\pm$0.01&0.82$\pm$0.01&0.83$\pm$0.01&0.83$\pm$0.01&0.64$\pm$0.03&0.59$\pm$0.07&0.68$\pm$0.03 \\ 
 & & & & & & & (+1 noise) &0.33$\pm$0.0&0.33$\pm$0.0&0.78$\pm$0.01&0.74$\pm$0.01&0.76$\pm$0.01&0.78$\pm$0.01&0.83$\pm$0.01&0.23$\pm$0.01&0.62$\pm$0.04&0.6$\pm$0.02\\
\inrule
\multirow{2}{*}{BBCSport}&\multirow{2}{*}{0.87}&\multirow{2}{*}{0.76}&\multirow{2}{*}{-}&\multirow{2}{*}{-}&\multirow{2}{*}{-}&\multirow{2}{*}{-}& & 0.63$\pm$0.07&0.63$\pm$0.07&0.87$\pm$0.03&0.87$\pm$0.03&0.87$\pm$0.03&0.87$\pm$0.03&0.87$\pm$0.03&0.73$\pm$0.09&0.79$\pm$0.09&0.7$\pm$0.06 \\
 & & & & & & & (+1 noise) &0.36$\pm$0.0&0.36$\pm$0.0&0.73$\pm$0.03&0.71$\pm$0.02&0.73$\pm$0.03&0.73$\pm$0.03&0.73$\pm$0.03&0.24$\pm$0.02&0.79$\pm$0.04&0.54$\pm$0.04\\
\inrule
\multirow{2}{*}{Wikipedia}&\multirow{2}{*}{0.17}&\multirow{2}{*}{0.61}&\multirow{2}{*}{-}&\multirow{2}{*}{-}&\multirow{2}{*}{-}&\multirow{2}{*}{-}& & 0.17$\pm$0.01&0.17$\pm$0.01&0.42$\pm$0.03&0.42$\pm$0.04&0.42$\pm$0.03&0.42$\pm$0.03&0.42$\pm$0.03&0.58$\pm$0.03&0.31$\pm$0.05&0.43$\pm$0.05 \\
 & & & & & & & (+1 noise) &0.15$\pm$0.0&0.15$\pm$0.0&0.34$\pm$0.01&0.34$\pm$0.01&0.34$\pm$0.01&0.34$\pm$0.01&0.33$\pm$0.02&0.15$\pm$0.0&0.26$\pm$0.07&0.33$\pm$0.03\\
\inrule
\multirow{2}{*}{UCI}&\multirow{2}{*}{0.79}&\multirow{2}{*}{0.72}&\multirow{2}{*}{0.9}&\multirow{2}{*}{0.43}&\multirow{2}{*}{0.91}&\multirow{2}{*}{0.69}& & 0.1$\pm$0.0&0.1$\pm$0.0&0.85$\pm$0.01&0.79$\pm$0.02&0.82$\pm$0.02&0.85$\pm$0.01&0.81$\pm$0.02&0.56$\pm$0.14&0.35$\pm$0.11&0.58$\pm$0.02 \\
 & & & & & &  & (+1 noise) &0.1$\pm$0.0&0.1$\pm$0.0&0.82$\pm$0.01&0.75$\pm$0.02&0.78$\pm$0.02&0.5$\pm$0.02&0.82$\pm$0.01&0.13$\pm$0.0&0.26$\pm$0.06&0.54$\pm$0.0\\
 \inrule
\multirow{2}{*}{cora}&\multirow{2}{*}{0.58}&\multirow{2}{*}{0.49}&\multirow{2}{*}{-}&\multirow{2}{*}{-}&\multirow{2}{*}{-}&\multirow{2}{*}{-}& & 0.3$\pm$0.01&0.3$\pm$0.01&0.56$\pm$0.04&0.56$\pm$0.04&0.56$\pm$0.04&0.57$\pm$0.03&0.56$\pm$0.03&0.55$\pm$0.05&0.37$\pm$0.07&0.45$\pm$0.03 \\ 
 & & & & & & &(+1 noise)&0.3$\pm$0.0&0.3$\pm$0.0&0.43$\pm$0.02&0.42$\pm$0.02&0.43$\pm$0.02&0.49$\pm$0.02&0.53$\pm$0.09&0.58$\pm$0.02&0.36$\pm$0.07&0.4$\pm$0.02\\
 \inrule
\multirow{2}{*}{citeseer}&\multirow{2}{*}{0.41}&\multirow{2}{*}{0.5}&\multirow{2}{*}{-}&\multirow{2}{*}{-}&\multirow{2}{*}{-}&\multirow{2}{*}{-}& & 0.23$\pm$0.01&0.23$\pm$0.01&0.53$\pm$0.04&0.54$\pm$0.03&0.54$\pm$0.04&0.55$\pm$0.03&0.54$\pm$0.03&0.4$\pm$0.01&0.3$\pm$0.08&0.38$\pm$0.03 \\ 
 & & & & & & &(+1 noise)&0.21$\pm$0.0&0.21$\pm$0.0&0.36$\pm$0.02&0.36$\pm$0.02&0.36$\pm$0.02&0.39$\pm$0.06&0.49$\pm$0.03&0.39$\pm$0.01&0.37$\pm$0.06&0.32$\pm$0.01\\
 \inrule
\multirow{2}{*}{dkpol}&\multirow{2}{*}{0.29}&\multirow{2}{*}{0.16}&\multirow{2}{*}{0.67}&\multirow{2}{*}{-}&\multirow{2}{*}{-}&\multirow{2}{*}{-}& & 0.15$\pm$0.0&0.15$\pm$0.0&0.67$\pm$0.03&0.66$\pm$0.03&0.67$\pm$0.03&0.67$\pm$0.03&0.67$\pm$0.03&0.19$\pm$0.04&0.24$\pm$0.06&0.57$\pm$0.02 \\ 
 & & & & & & &(+1 noise)&0.14$\pm$0.0&0.14$\pm$0.0&0.57$\pm$0.04&0.55$\pm$0.03&0.56$\pm$0.03&0.65$\pm$0.04&0.65$\pm$0.04&0.17$\pm$0.01&0.34$\pm$0.15&0.31$\pm$0.02\\
 \inrule
\multirow{2}{*}{aucs}&\multirow{2}{*}{0.34}&\multirow{2}{*}{0.36}&\multirow{2}{*}{0.61}&\multirow{2}{*}{0.8}&\multirow{2}{*}{0.72}&\multirow{2}{*}{-}& & 0.3$\pm$0.01&0.3$\pm$0.01&0.85$\pm$0.04&0.8$\pm$0.06&0.85$\pm$0.05&0.84$\pm$0.04&0.85$\pm$0.03&0.75$\pm$0.07&0.5$\pm$0.16&0.81$\pm$0.05 \\ 
 & & & & & & &(+1 noise)&0.27$\pm$0.0&0.27$\pm$0.0&0.81$\pm$0.04&0.37$\pm$0.03&0.65$\pm$0.03&0.81$\pm$0.04&0.81$\pm$0.04&0.76$\pm$0.09&0.42$\pm$0.03&0.77$\pm$0.09\\
 \midrule
 APR & & & & & & & 
 &0.41&0.41&0.94&0.87&0.91&0.94&\textbf{0.97}&0.66&0.65&0.77\\
  AR & & & & & & & 
 &9.1&9.1&3.0&5.1&3.9&2.8&\textbf{2.5}&6.5&6.8&6.1\\
\bottomrule
\end{NiceTabular}}
\end{scriptsize}
\end{table*}

\def\inrule{ \specialrule{.01pt}{1pt}{1pt} }
\setlength{\tabcolsep}{2.5pt}
 \begin{table*}[t]
\caption{Accuracy (mean $\pm$ std) over three random samples of the 10\% of input labels, on real-world datasets (+ one layer of noise).}
\label{tab:1layer10percent}
\centering
\begin{scriptsize}
\resizebox{\textwidth}{!}{
\begin{NiceTabular}{c|cccccc|cccccc|cc|ccc}
\toprule
 & \RNum{1} & \RNum{2} & \RNum{3} & \RNum{4} & \RNum{5} & \RNum{6} &  & MIN & GEOM & ARIT & HARM & MAX & BINOM & MULTI & SGMI & SMACD & GMM\\
\midrule
\multirow{2}{*}{3sources}&\multirow{2}{*}{0.76}&\multirow{2}{*}{0.72}&\multirow{2}{*}{0.75}&\multirow{2}{*}{-}&\multirow{2}{*}{-}&\multirow{2}{*}{-}& & 0.71$\pm$0.07&0.71$\pm$0.07&0.79$\pm$0.07&0.69$\pm$0.05&0.75$\pm$0.05&0.75$\pm$0.07&0.74$\pm$0.05&0.75$\pm$0.04&0.62$\pm$0.04&0.8$\pm$0.05 \\
 & & & & & & & (+1 noise) &0.34$\pm$0.01&0.34$\pm$0.01&0.74$\pm$0.1&0.54$\pm$0.06&0.66$\pm$0.09&0.76$\pm$0.11&0.73$\pm$0.06&0.58$\pm$0.27&0.6$\pm$0.08&0.76$\pm$0.04\\
 \inrule
\multirow{2}{*}{BBC}&\multirow{2}{*}{0.83}&\multirow{2}{*}{0.83}&\multirow{2}{*}{0.79}&\multirow{2}{*}{0.83}&\multirow{2}{*}{-}&\multirow{2}{*}{-}& & 0.38$\pm$0.01&0.38$\pm$0.01&0.91$\pm$0.01&0.89$\pm$0.01&0.9$\pm$0.01&0.88$\pm$0.01&0.86$\pm$0.04&0.76$\pm$0.02&0.69$\pm$0.03&0.87$\pm$0.01 \\ 
 & & & & & & & (+1 noise) &0.33$\pm$0.0&0.33$\pm$0.0&0.89$\pm$0.01&0.84$\pm$0.01&0.87$\pm$0.02&0.87$\pm$0.02&0.87$\pm$0.01&0.76$\pm$0.03&0.66$\pm$0.02&0.84$\pm$0.01\\
\inrule
\multirow{2}{*}{BBCSport}&\multirow{2}{*}{0.90}&\multirow{2}{*}{0.88}&\multirow{2}{*}{-}&\multirow{2}{*}{-}&\multirow{2}{*}{-}&\multirow{2}{*}{-}& & 0.79$\pm$0.0&0.79$\pm$0.0&0.92$\pm$0.01&0.92$\pm$0.0&0.92$\pm$0.0&0.92$\pm$0.02&0.88$\pm$0.01&0.84$\pm$0.03&0.73$\pm$0.1&0.88$\pm$0.02 \\
 & & & & & & & (+1 noise) &0.36$\pm$0.0&0.36$\pm$0.0&0.86$\pm$0.02&0.82$\pm$0.01&0.85$\pm$0.02&0.91$\pm$0.01&0.87$\pm$0.0&0.83$\pm$0.04&0.6$\pm$0.07&0.77$\pm$0.02\\
\inrule
\multirow{2}{*}{Wikipedia}&\multirow{2}{*}{0.18}&\multirow{2}{*}{0.65}&\multirow{2}{*}{-}&\multirow{2}{*}{-}&\multirow{2}{*}{-}&\multirow{2}{*}{-}& & 0.19$\pm$0.01&0.19$\pm$0.01&0.51$\pm$0.03&0.51$\pm$0.03&0.51$\pm$0.03&0.62$\pm$0.04&0.64$\pm$0.02&0.61$\pm$0.03&0.24$\pm$0.04&0.57$\pm$0.03 \\
 & & & & & & & (+1 noise) &0.15$\pm$0.0&0.15$\pm$0.0&0.42$\pm$0.01&0.42$\pm$0.01&0.42$\pm$0.01&0.57$\pm$0.06&0.62$\pm$0.03&0.59$\pm$0.02&0.25$\pm$0.04&0.47$\pm$0.02\\
\inrule
\multirow{2}{*}{UCI}&\multirow{2}{*}{0.92}&\multirow{2}{*}{0.81}&\multirow{2}{*}{0.96}&\multirow{2}{*}{0.57}&\multirow{2}{*}{0.97}&\multirow{2}{*}{0.82}& & 0.11$\pm$0.0&0.11$\pm$0.0&0.95$\pm$0.01&0.88$\pm$0.01&0.92$\pm$0.01&0.97$\pm$0.01&0.96$\pm$0.01&0.94$\pm$0.01&0.33$\pm$0.05&0.93$\pm$0.01 \\
 & & & & & &  & (+1 noise) &0.1$\pm$0.0&0.1$\pm$0.0&0.96$\pm$0.0&0.88$\pm$0.01&0.92$\pm$0.0&0.97$\pm$0.0&0.96$\pm$0.01&0.94$\pm$0.01&0.3$\pm$0.1&0.93$\pm$0.01\\
 \inrule
\multirow{2}{*}{cora}&\multirow{2}{*}{0.73}&\multirow{2}{*}{0.62}&\multirow{2}{*}{-}&\multirow{2}{*}{-}&\multirow{2}{*}{-}&\multirow{2}{*}{-}& & 0.34$\pm$0.01&0.34$\pm$0.01&0.69$\pm$0.01&0.69$\pm$0.01&0.69$\pm$0.01&0.74$\pm$0.03&0.76$\pm$0.01&0.72$\pm$0.01&0.34$\pm$0.08&0.69$\pm$0.01 \\ 
 & & & & & & &(+1 noise)&0.3$\pm$0.0&0.3$\pm$0.0&0.57$\pm$0.01&0.56$\pm$0.01&0.57$\pm$0.01&0.63$\pm$0.01&0.76$\pm$0.01&0.72$\pm$0.02&0.28$\pm$0.01&0.58$\pm$0.01\\
 \inrule
\multirow{2}{*}{citeseer}&\multirow{2}{*}{0.51}&\multirow{2}{*}{0.62}&\multirow{2}{*}{-}&\multirow{2}{*}{-}&\multirow{2}{*}{-}&\multirow{2}{*}{-}& & 0.29$\pm$0.01&0.29$\pm$0.01&0.65$\pm$0.01&0.65$\pm$0.01&0.65$\pm$0.01&0.66$\pm$0.01&0.65$\pm$0.01&0.51$\pm$0.01&0.36$\pm$0.1&0.58$\pm$0.01 \\ 
 & & & & & & &(+1 noise)&0.21$\pm$0.0&0.21$\pm$0.0&0.53$\pm$0.02&0.52$\pm$0.02&0.53$\pm$0.02&0.59$\pm$0.03&0.63$\pm$0.0&0.52$\pm$0.02&0.32$\pm$0.03&0.49$\pm$0.01\\
 \inrule
\multirow{2}{*}{dkpol}&\multirow{2}{*}{0.32}&\multirow{2}{*}{0.16}&\multirow{2}{*}{0.69}&\multirow{2}{*}{-}&\multirow{2}{*}{-}&\multirow{2}{*}{-}& & 0.16$\pm$0.0&0.16$\pm$0.0&0.73$\pm$0.06&0.67$\pm$0.09&0.69$\pm$0.07&0.62$\pm$0.06&0.76$\pm$0.01&0.31$\pm$0.03&0.26$\pm$0.09&0.63$\pm$0.04 \\ 
 & & & & & & &(+1 noise)&0.14$\pm$0.0&0.14$\pm$0.0&0.62$\pm$0.02&0.59$\pm$0.02&0.6$\pm$0.02&0.54$\pm$0.08&0.72$\pm$0.08&0.31$\pm$0.05&0.24$\pm$0.01&0.34$\pm$0.01\\
 \inrule
\multirow{2}{*}{aucs}&\multirow{2}{*}{0.34}&\multirow{2}{*}{0.36}&\multirow{2}{*}{0.61}&\multirow{2}{*}{0.8}&\multirow{2}{*}{0.72}&\multirow{2}{*}{-}& & 0.3$\pm$0.01&0.3$\pm$0.01&0.85$\pm$0.04&0.8$\pm$0.06&0.85$\pm$0.05&0.85$\pm$0.04&0.85$\pm$0.02&0.75$\pm$0.07&0.58$\pm$0.03&0.81$\pm$0.05 \\ 
 & & & & & & &(+1 noise)&0.27$\pm$0.0&0.27$\pm$0.0&0.81$\pm$0.04&0.37$\pm$0.03&0.65$\pm$0.03&0.81$\pm$0.04&0.81$\pm$0.04&0.76$\pm$0.09&0.56$\pm$0.07&0.77$\pm$0.09\\
 \midrule
 APR & & & & & & & 
 &0.38&0.38&0.93&0.85&0.89&0.95&\textbf{0.98}&0.85&0.55&0.88\\
  AR & & & & & & & 
 &9.2&9.2&3.1&5.7&4.4&2.6&\textbf{2.4}&5.1&8.1&5.0\\
\bottomrule
\end{NiceTabular}}
\end{scriptsize}
\end{table*}

\bibliography{biblio}

\newpage
\onecolumngrid

\appendix
\section{Proof of \cref{nonconvb}} \label{AppendixA}
The following chain of inequalities holds:
\begin{equation} \label{chain}
    -\nabla f(\boldsymbol{\theta}_n)^\top d_n^{FW} \geq
    -\nabla f(\boldsymbol{\theta}_n)^\top d_n \geq
    -\widetilde \nabla f(\boldsymbol{\theta}_n)^\top d_n - \epsilon_n \geq
    -\widetilde \nabla f(\boldsymbol{\theta}_n)^\top d_n^{FW} - \epsilon_n \geq
    -\nabla f(\boldsymbol{\theta}_n)^\top d_n^{FW} - 2\epsilon_n,
\end{equation}
where we used \eqref{cond_inexact} in the second and the last inequality, while the first and the third inequality follow from the definition of $d_n^{FW}$ and $d_n$.
In particular, using the definitions of $\tilde g^n$, $g_n$ and $g_n^{FW}$, from~\eqref{chain} we can write
\begin{align}
g_n^{FW} & \ge \tilde g_n - \epsilon_n, \label{ineq_proof_1} \\
\tilde g_n & \ge g_n^{FW} - \epsilon_n, \label{ineq_proof_2} \\
g_n & \ge \tilde g_n - \epsilon_n \label{ineq_proof_3}
 \end{align}

Using~\eqref{assumption} and~\eqref{ineq_proof_1}, we also have
\begin{equation}\label{chain_2}
\epsilon_n \leq \sigma (\tilde g_n - \epsilon_n) \leq \sigma g_n^{FW},
\end{equation}
Now, let us distinguish two cases.
\begin{itemize}
\item 
If $\bar{\eta}_n < 1$, from~\eqref{alphabound} it follows that $\displaystyle{\frac{\tilde{g}_n}{{M\n{d_n}^2}}<1}$. Using~\eqref{eq:rho} we can write
	\begin{equation*}
	f(\boldsymbol{\theta}_n) - f(\boldsymbol{\theta}_n + \eta_n d_n) \ge \rho\bar{\eta}_n\tilde{g}_n = \frac{\rho}{M{\n{d_n}}^2} \tilde{g}_n^2 \ge \frac{\rho \tilde{g}_n^2} {\Delta^2M},
	\end{equation*}
    where the last inequality follows from $\n{d_n} \leq \Delta$.
    Observe that, from~\eqref{ineq_proof_2} and~\eqref{chain_2}, we have
    $\tilde g_n \ge (1-\sigma) g_n^{FW}$.
    Therefore,
        \begin{equation} \label{c1}
	f(\boldsymbol{\theta}_n) - f(\boldsymbol{\theta}_{n+1}) \ge 
        \frac{\rho (1 - \sigma)^2} {\Delta^2M} (g_n^{FW})^2.
	\end{equation}	

\item If $\bar{\eta}_n = 1$, from~\eqref{alphabound} it follows that $\displaystyle{\frac{\tilde{g}_n}{{M\n{d_n}^2}}\ge 1}$ and, since $\eta_n \le 1$ from the instructions of the algorithm, then $\eta_n = 1$. 
	By the standard descent lemma we can write
	\[ 
 f(\boldsymbol{\theta}_{n+1}) = f(\boldsymbol{\theta}_n + d_n) \leq f(\boldsymbol{\theta}_n) - g_n + \frac{M}{2}\|d_n\|^2\,   
        \leq f(\boldsymbol{\theta}_n) - (\tilde{g}_n - \epsilon_n) + \frac{M}{2}\|d_n\|^2\, ,
        \]
        where we used \eqref{ineq_proof_3} in the last inequality.
	Since we are analyzing the case where $\tilde{g}_n \ge \n{d_n}^2M$, we obtain
    \[
    f(\boldsymbol{\theta}_n) - f(\boldsymbol{\theta}_{n+1}) \ge \frac{\tilde{g}_n}{2} - \epsilon_n.
    \]
    Using~\eqref{ineq_proof_2} and~\eqref{chain_2}, we also have
    \[
    \frac{\tilde{g}_n}{2} - \epsilon_n \ge \frac{g_n^{FW}}{2} - \frac{3}{2}\epsilon_n \ge 
 \frac{g_n^{FW}}{2} - \frac{3}{2}\sigma g_n^{FW} =
 \frac{1 - 3 \sigma}{2} g_n^{FW}.
    \]
    Therefore,
    \begin{equation} \label{c2}
	f(\boldsymbol{\theta}_n) - f(\boldsymbol{\theta}_{n+1}) \ge
 \frac{1 - 3 \sigma}{2} g_n^{FW}.
	\end{equation}
 \end{itemize}

 Now, based on the two cases analyzed above, we partition the iterations $\{0,1,\ldots,T-1\}$ into two subsets $N_1$ and $N_2$ defined as follows:
\[
N_1 = \{n < T \colon \bar{\eta}_n < 1\}, \quad
N_2 = \{n < T \colon \bar{\eta}_n = 1\}.
\]

	Using~\eqref{c1} and~\eqref{c2}, we can write:
        \[
        \begin{split}
        f(\boldsymbol{\theta}_0)-f^*  &\ge \sum_{n=0}^{T-1} (f(\boldsymbol{\theta}_n)-f(\boldsymbol{\theta}_{n+1})) \\
        & = \sum_{N_1} (f(\boldsymbol{\theta}_n)-f(\boldsymbol{\theta}_{n+1}))+
        \sum_{N_2} (f(\boldsymbol{\theta}_n)-f(\boldsymbol{\theta}_{n+1})) \\
        & \ge \sum_{N_1} \frac{\rho (1 - \sigma)^2} {\Delta^2M}(g_n^{FW})^2 +
        \sum_{N_2} \frac{1 - 3 \sigma}{2} g_n^{FW} \\
        & \ge |N_1| \min_{n \in N_1} \frac{\rho (1 - \sigma)^2} {\Delta^2M}(g_n^{FW})^2 + |N_2| \min_{n \in N_2} \frac{1 - 3 \sigma}{2} g_n^{FW} \\
        & \ge (|N_1|+|N_2|) \min \left(\frac{\rho (1 - \sigma)^2} {\Delta^2M}(g_T^{*})^2 ,\frac{1 - 3 \sigma}{2} g_T^{*} \right) \\
        & = T \min \left(\frac{\rho (1 - \sigma)^2} {\Delta^2M}(g_T^{*})^2 ,\frac{1 - 3 \sigma}{2} g_T^{*} \right),
        \end{split}
        \]
        where the last inequality follows from the definition of $g_T^{*}$.
    Hence,
    \begin{align*}
    \frac{\rho (1 - \sigma)^2} {\Delta^2M}(g_T^{*})^2 \le \frac{1 - 3 \sigma}{2} g_T^{*} \; & \Rightarrow  \;
   g^*_{T} \leq \sqrt{\frac{\Delta^2 M (f(\boldsymbol{\theta}_0) - f^*)}{T \rho ( 1 - \sigma)^2}}, \\
    \frac{\rho (1 - \sigma)^2} {\Delta^2M}(g_T^{*})^2 > \frac{1 - 3 \sigma}{2} g_T^{*} \; & \Rightarrow  \;
    g^*_{T} \le \frac{2(f(\boldsymbol{\theta}_0) - f^*)}{T(1 - 3\sigma)},
    \end{align*}
    leading to the desired result. 
\section{Proof of \cref{lemma}}  \label{AppendixB}
Reasoning as in the proof of \cref{nonconvb}, we have that~\eqref{ineq_proof_3} holds.
By the standard descent lemma, we have
\begin{equation} \label{e11:std}
f(\boldsymbol{\theta}_n) - f(\boldsymbol{\theta}_n + \eta d_n) \ge \eta g_n - \eta^2 \frac{M\|d_n\|^2}{2}
\ge \eta (\tilde{g}_n - \epsilon_n) - \eta^2 \frac{M\|d_n\|^2}{2}, \quad \forall \eta \in \mathbb R,
\end{equation}
where the last inequality follows from~\eqref{ineq_proof_3}.
Then,
\[
f(\boldsymbol{\theta}_n) - f(\boldsymbol{\theta}_n + \eta d_n) \ge \gamma \eta \tilde{g}_n \quad \forall \eta \in \left[0, 2\frac{(1 - \gamma) \tilde{g}_n- \epsilon_n}{M\n{d_n}^2} \right].
\]

Since $\eta_n$ is computed by~\eqref{eta}--\eqref{Armijo}, we can write
	\begin{equation} \label{1Ar}
        \begin{split} 
\eta_n 
  & \ge \min\left(1,2\delta\frac{(1 - \gamma) \tilde{g}_n- \epsilon_n}
 {M\n{d_n}^2} \right)\\
 & \ge \min\left(1,2\delta\frac{(1 - \gamma -\sigma) \tilde{g}_n}
 {M\n{d_n}^2} \right)\\
  & \ge \min(1,2\delta (1- \gamma -\sigma))\bar \eta_n,
        \end{split} 
	\end{equation}
 where the second inequality follows from~\eqref{assumption2}. 

\clearpage
\section{Complexity Analysis} \label{AppendixD}
We detail below the computational cost of the method proposed in Section \ref{Theory} and provide a table with a time-execution comparison with respect to the competing methods from Table~\ref{tab:info2} and Table~\ref{tab:1layer1percent}-\ref{tab:1layer10percent}.

In \cref{nonconvb} we have shown a sublinear convergence rate of the duality gap $g_{n}^{FW}$, that is, $g_ n^* \leq \max(c_1 n^{-\frac{1}{2}}, c_2 n^{-1})$ with appropriate constants $c_1$ and $c_2$. Then, complexity results can be straightforwardly obtained by standard arguments of information-based complexity theory 
\cite{nesterov2003introductory}. In particular, in our case we have a worst-case complexity of $\mathcal{O}(\epsilon^{-2})$ for the number of iterations to drive $g_ n^*$ below $\epsilon$. Additionally, we can easily give an upper bound on the number of arithmetic operations carried out at every iteration of the FW method: each iteration requires $\mathcal{O}(K)$ function evaluations to estimate the gradient and $\mathcal{O}(K)$ operations to solve the linear subproblem (the line search has a cost $\mathcal{O}(1)$ assuming the knowledge of the Lipschitz constant $M$), where $K$ is the number of layers. Moreover, each function evaluation requires the solution of problem \eqref{bprob} (or problem \eqref{bprob_single}), that is, $r$ iterations of Label Propagation algorithm yielding a cost of $\mathcal{O}(Nr)$ if the graph is sparse, where N is the number of nodes. Summing up, each iteration of the FW method a cost of $\mathcal{O}(KNr)$ and then we need $\mathcal{O}(\epsilon^{-2}KNr)$ arithmetic operations to drive $g_ n^*$ below $\epsilon$.\\
In Table~\ref{tab:time-execution}, we report the average time-execution comparison over three runs (in seconds) on synthetic datasets generated as in Subsection~\ref{Synthetic Datasets} (3 communities equal size and 3 layers) as the number of nodes increases.

\setlength{\tabcolsep}{2.5pt}
 \begin{table*}[!h]
\caption{
Average time-execution comparison over three runs (in seconds) on synthetic datasets with 3 communities of equal size and 3 layers, as the number of nodes $N$ increases.}
\label{tab:time-execution}
\vskip 0.15in
\centering
\begin{scriptsize}
\begin{NiceTabular}{c|ccc|ccccc|cc|cccc}
    \toprule
        N & 1 & 2 & 3 &  MIN & GEOM & ARIT & HARM & MAX & BINOM & MULTI & SGMI & AGML & SMACD & GMM \\ 
        \midrule
        
       1200 & 0.007 & 0.007 & 0.007 & 0.011 & 0.012 & 0.009 & 0.005 & 0.005 & 6.1 & 15.017 & 0.155 & 0.711 & 26.397 & 1.027 \\
    2400 & 0.011 & 0.01 & 0.01 & 0.017 & 0.018 & 0.014 & 0.005 & 0.005 & 6.493 & 17.614 & 0.72 & 5.434 & 58.344 & 9.136 \\
    3600 & 0.014 & 0.013 & 0.012 & 0.021 & 0.022 & 0.017 & 0.005 & 0.006 & 6.914 & 17.143 & 1.47 & 13.608 & 97.081 & 27.543 \\
    4800 & 0.016 & 0.014 & 0.014 & 0.024 & 0.026 & 0.022 & 0.006 & 0.007 & 9.675 & 17.973 & 3.132 & 31.458 & 127.851 & 73.991 \\
    6000 & 0.018 & 0.016 & 0.016 & 0.028 & 0.032 & 0.028 & 0.008 & 0.008 & 9.074 & 28.186 & 6.809 & 60.876 & 132.44 & 173.659 \\
    7200 & 0.02 & 0.018 & 0.017 & 0.033 & 0.037 & 0.033 & 0.009 & 0.009 & 9.98 & 26.953 & 10.618 & 123.825 & 188.622 & 299.11 \\
    8400 & 0.024 & 0.021 & 0.02 & 0.04 & 0.045 & 0.04 & 0.01 & 0.01 & 11.984 & 29.765 & 10.586 & 132.112 & 157.038 & 488.309 \\
    9600 & 0.025 & 0.023 & 0.019 & 0.038 & 0.041 & 0.037 & 0.009 & 0.009 & 12.959 & 32.789 & 21.794 & 294.769 & 285.303 & 789.337 \\
    10800 & 0.034 & 0.027 & 0.023 & 0.044 & 0.173 & 0.045 & 0.011 & 0.011 & 32.9 & 98.419 & 47.944 & 561.55 & 348.047 & 1040.754 \\
    12000 & 0.026 & 0.023 & 0.022 & 0.047 & 0.051 & 0.047 & 0.011 & 0.011 & 30.391 & 96.565 & 44.969 & 546.674 & 393.805 & 1626.726 \\ 
    \bottomrule    
    \end{NiceTabular}
\vskip -0.1in
\end{scriptsize}
\end{table*}

\section{Additional Results}  \label{sec:additional_results}

In Tables~\ref{tab:2layers1percent}-\ref{tab:2layers15percent}, we report tests performed with different number of initial known labels per community (1\%, 5\%, 10\%, 15\%), considering the informative case and the adding of one or two layers of noise. 
The results are aligned with those presented in Tables~\ref{tab:1layer1percent} and \ref{tab:1layer10percent}, with the proposed BINOM and MULTI approaches being overall the best performing.

 \begin{table*}[h!]
\caption{Accuracy (mean $\pm$ std) over three random samples of the 1\% of input labels, on real-world datasets (+ one and two layers of noise).}
\label{tab:2layers1percent}
\centering
\begin{scriptsize}
\resizebox{\textwidth}{!}{
\begin{NiceTabular}{c|cccccc|cccccc|cc|ccc}
\toprule
 & \RNum{1} & \RNum{2} & \RNum{3} & \RNum{4} & \RNum{5} & \RNum{6} &  & MIN & GEOM & ARIT & HARM & MAX & BINOM & MULTI & SGMI & SMACD & GMM\\
\midrule
\multirow{3}{*}{3sources}&\multirow{3}{*}{0.72}&\multirow{3}{*}{0.69}&\multirow{3}{*}{0.77}&\multirow{3}{*}{-}&\multirow{3}{*}{-}&\multirow{3}{*}{-}& & 0.66$\pm$0.01&0.66$\pm$0.01&0.79$\pm$0.04&0.71$\pm$0.04&0.76$\pm$0.04&0.79$\pm$0.04&0.79$\pm$0.04&0.65$\pm$0.04&0.58$\pm$0.06&0.69$\pm$0.06 \\
 & & & & & & & (+1 noise) &0.35$\pm$0.01&0.35$\pm$0.01&0.73$\pm$0.06&0.55$\pm$0.05&0.69$\pm$0.07&0.73$\pm$0.06&0.73$\pm$0.06&0.52$\pm$0.22&0.65$\pm$0.08&0.62$\pm$0.02\\
 & & & & & & & (+2 noise) &0.34$\pm$0.0&0.34$\pm$0.0&0.7$\pm$0.04&0.48$\pm$0.02&0.63$\pm$0.07&0.7$\pm$0.04&0.7$\pm$0.04&0.52$\pm$0.22&0.55$\pm$0.07&0.53$\pm$0.03\\
 \inrule
\multirow{3}{*}{BBC}&\multirow{3}{*}{0.77}&\multirow{3}{*}{0.76}&\multirow{3}{*}{0.72}&\multirow{3}{*}{0.75}&\multirow{3}{*}{-}&\multirow{3}{*}{-}& & 0.35$\pm$0.01&0.35$\pm$0.01&0.83$\pm$0.01&0.81$\pm$0.01&0.82$\pm$0.01&0.83$\pm$0.01&0.83$\pm$0.01&0.64$\pm$0.03&0.59$\pm$0.07&0.68$\pm$0.03 \\ 
 & & & & & & & (+1 noise) &0.33$\pm$0.0&0.33$\pm$0.0&0.78$\pm$0.01&0.74$\pm$0.01&0.76$\pm$0.01&0.78$\pm$0.01&0.83$\pm$0.01&0.23$\pm$0.01&0.62$\pm$0.04&0.6$\pm$0.02\\
  & & & & & & & (+2 noise) & 0.33$\pm$0.0&0.33$\pm$0.0&0.73$\pm$0.01&0.68$\pm$0.02&0.71$\pm$0.01&0.73$\pm$0.01&0.73$\pm$0.01&0.23$\pm$0.01&0.56$\pm$0.04&0.55$\pm$0.02\\
\inrule
\multirow{3}{*}{BBCSport}&\multirow{3}{*}{0.87}&\multirow{3}{*}{0.76}&\multirow{3}{*}{-}&\multirow{3}{*}{-}&\multirow{3}{*}{-}&\multirow{3}{*}{-}& & 0.63$\pm$0.07&0.63$\pm$0.07&0.87$\pm$0.03&0.87$\pm$0.03&0.87$\pm$0.03&0.87$\pm$0.03&0.87$\pm$0.03&0.73$\pm$0.09&0.79$\pm$0.09&0.7$\pm$0.06 \\
 & & & & & & & (+1 noise) &0.36$\pm$0.0&0.36$\pm$0.0&0.73$\pm$0.03&0.71$\pm$0.02&0.73$\pm$0.03&0.73$\pm$0.03&0.73$\pm$0.03&0.24$\pm$0.02&0.79$\pm$0.04&0.54$\pm$0.04\\
  & & & & & & & (+2 noise) & 0.36$\pm$0.0&0.36$\pm$0.0&0.66$\pm$0.03&0.62$\pm$0.03&0.65$\pm$0.03&0.66$\pm$0.03&0.77$\pm$0.07&0.24$\pm$0.02&0.73$\pm$0.16&0.5$\pm$0.04\\
\inrule
\multirow{3}{*}{Wikipedia}&\multirow{3}{*}{0.17}&\multirow{3}{*}{0.61}&\multirow{3}{*}{-}&\multirow{3}{*}{-}&\multirow{3}{*}{-}&\multirow{3}{*}{-}& & 0.17$\pm$0.01&0.17$\pm$0.01&0.42$\pm$0.03&0.42$\pm$0.04&0.42$\pm$0.03&0.42$\pm$0.03&0.42$\pm$0.03&0.58$\pm$0.03&0.31$\pm$0.05&0.43$\pm$0.05 \\
 & & & & & & & (+1 noise) &0.15$\pm$0.0&0.15$\pm$0.0&0.34$\pm$0.01&0.34$\pm$0.01&0.34$\pm$0.01&0.34$\pm$0.01&0.33$\pm$0.02&0.15$\pm$0.0&0.26$\pm$0.07&0.33$\pm$0.03\\
  & & & & & & & (+2 noise) &0.15$\pm$0.0&0.15$\pm$0.0&0.32$\pm$0.01&0.31$\pm$0.02&0.31$\pm$0.02&0.32$\pm$0.01&0.32$\pm$0.01&0.15$\pm$0.0&0.23$\pm$0.08&0.28$\pm$0.03\\
\inrule
\multirow{3}{*}{UCI}&\multirow{3}{*}{0.79}&\multirow{3}{*}{0.72}&\multirow{3}{*}{0.9}&\multirow{3}{*}{0.43}&\multirow{3}{*}{0.91}&\multirow{3}{*}{0.69}& & 0.1$\pm$0.0&0.1$\pm$0.0&0.85$\pm$0.01&0.79$\pm$0.02&0.82$\pm$0.02&0.85$\pm$0.01&0.81$\pm$0.02&0.56$\pm$0.14&0.35$\pm$0.11&0.58$\pm$0.02 \\
 & & & & & &  & (+1 noise) &0.1$\pm$0.0&0.1$\pm$0.0&0.82$\pm$0.01&0.75$\pm$0.02&0.78$\pm$0.02&0.5$\pm$0.02&0.82$\pm$0.01&0.13$\pm$0.0&0.26$\pm$0.06&0.54$\pm$0.0\\
  & & & & & & & (+2 noise) & 0.1$\pm$0.0&0.1$\pm$0.0&0.79$\pm$0.01&0.71$\pm$0.02&0.75$\pm$0.02&0.89$\pm$0.03&0.89$\pm$0.03&0.13$\pm$0.0&0.23$\pm$0.06&0.51$\pm$0.01\\
 \inrule
\multirow{3}{*}{cora}&\multirow{3}{*}{0.58}&\multirow{3}{*}{0.49}&\multirow{3}{*}{-}&\multirow{3}{*}{-}&\multirow{3}{*}{-}&\multirow{3}{*}{-}& & 0.3$\pm$0.01&0.3$\pm$0.01&0.56$\pm$0.04&0.56$\pm$0.04&0.56$\pm$0.04&0.57$\pm$0.03&0.56$\pm$0.03&0.55$\pm$0.05&0.37$\pm$0.07&0.45$\pm$0.03 \\ 
 & & & & & & &(+1 noise)&0.3$\pm$0.0&0.3$\pm$0.0&0.43$\pm$0.02&0.42$\pm$0.02&0.43$\pm$0.02&0.49$\pm$0.02&0.53$\pm$0.09&0.58$\pm$0.02&0.36$\pm$0.07&0.4$\pm$0.02\\
  & & & & & & & (+2 noise) & 0.3$\pm$0.0&0.3$\pm$0.0&0.35$\pm$0.02&0.34$\pm$0.02&0.34$\pm$0.02&0.44$\pm$0.04&0.53$\pm$0.09&0.58$\pm$0.02&0.36$\pm$0.05&0.36$\pm$0.01\\
 \inrule
\multirow{3}{*}{citeseer}&\multirow{3}{*}{0.41}&\multirow{3}{*}{0.5}&\multirow{3}{*}{-}&\multirow{3}{*}{-}&\multirow{3}{*}{-}&\multirow{3}{*}{-}& & 0.23$\pm$0.01&0.23$\pm$0.01&0.53$\pm$0.04&0.54$\pm$0.03&0.54$\pm$0.04&0.55$\pm$0.03&0.54$\pm$0.03&0.4$\pm$0.01&0.3$\pm$0.08&0.38$\pm$0.03 \\ 
 & & & & & & &(+1 noise)&0.21$\pm$0.0&0.21$\pm$0.0&0.36$\pm$0.02&0.36$\pm$0.02&0.36$\pm$0.02&0.39$\pm$0.06&0.49$\pm$0.03&0.39$\pm$0.01&0.37$\pm$0.06&0.32$\pm$0.01\\
  & & & & & & & (+2 noise) &0.21$\pm$0.0&0.21$\pm$0.0&0.3$\pm$0.02&0.3$\pm$0.02&0.3$\pm$0.02&0.32$\pm$0.04&0.5$\pm$0.02&0.39$\pm$0.01&0.25$\pm$0.05&0.29$\pm$0.0\\
 \inrule
\multirow{3}{*}{dkpol}&\multirow{3}{*}{0.29}&\multirow{3}{*}{0.16}&\multirow{3}{*}{0.67}&\multirow{3}{*}{-}&\multirow{3}{*}{-}&\multirow{3}{*}{-}& & 0.15$\pm$0.0&0.15$\pm$0.0&0.67$\pm$0.03&0.66$\pm$0.03&0.67$\pm$0.03&0.67$\pm$0.03&0.67$\pm$0.03&0.19$\pm$0.04&0.24$\pm$0.06&0.57$\pm$0.02 \\ 
 & & & & & & &(+1 noise)&0.14$\pm$0.0&0.14$\pm$0.0&0.57$\pm$0.04&0.55$\pm$0.03&0.56$\pm$0.03&0.65$\pm$0.04&0.65$\pm$0.04&0.17$\pm$0.01&0.34$\pm$0.15&0.31$\pm$0.02\\
  & & & & & & & (+2 noise) &0.14$\pm$0.0&0.14$\pm$0.0&0.51$\pm$0.03&0.49$\pm$0.03&0.5$\pm$0.03&0.33$\pm$0.02&0.33$\pm$0.02&0.17$\pm$0.01&0.22$\pm$0.03&0.23$\pm$0.04\\
 \inrule
\multirow{3}{*}{aucs}&\multirow{3}{*}{0.34}&\multirow{3}{*}{0.36}&\multirow{3}{*}{0.61}&\multirow{3}{*}{0.8}&\multirow{3}{*}{0.72}&\multirow{3}{*}{-}& & 0.3$\pm$0.01&0.3$\pm$0.01&0.85$\pm$0.04&0.8$\pm$0.06&0.85$\pm$0.05&0.84$\pm$0.04&0.85$\pm$0.03&0.75$\pm$0.07&0.5$\pm$0.16&0.81$\pm$0.05 \\ 
 & & & & & & &(+1 noise)&0.27$\pm$0.0&0.27$\pm$0.0&0.81$\pm$0.04&0.37$\pm$0.03&0.65$\pm$0.03&0.81$\pm$0.04&0.81$\pm$0.04&0.76$\pm$0.09&0.42$\pm$0.03&0.77$\pm$0.09\\
  & & & & & & & (+2 noise) & 0.27$\pm$0.0&0.27$\pm$0.0&0.77$\pm$0.05&0.33$\pm$0.02&0.52$\pm$0.04&0.73$\pm$0.1&0.73$\pm$0.1&0.76$\pm$0.09&0.52$\pm$0.08&0.75$\pm$0.09\\
 \midrule
 APR & & & & & & & 
 &0.4&0.4&0.92&0.83&0.88&0.92&\textbf{0.96}&0.62&0.64&0.74\\
  AR & & & & & & & 
 & 9.1&9.1&2.9&5.4&4.1&2.8& \textbf{2.4}&6.5&6.6&6.0\\
\bottomrule
\end{NiceTabular}}
\end{scriptsize}
\end{table*}

 \begin{table*}[h!]
\caption{Accuracy (mean $\pm$ std) over three random samples of the 5\% of input labels, on real-world datasets (+ one and two layers of noise).}
\label{tab:2layers5percent}
\centering
\begin{scriptsize}
\resizebox{\textwidth}{!}{
\begin{NiceTabular}{c|cccccc|cccccc|cc|ccc}
\toprule
 & \RNum{1} & \RNum{2} & \RNum{3} & \RNum{4} & \RNum{5} & \RNum{6} &  & MIN & GEOM & ARIT & HARM & MAX & BINOM & MULTI & SGMI & SMACD & GMM\\
\midrule
\multirow{3}{*}{3sources}&\multirow{3}{*}{0.73}&\multirow{3}{*}{0.72}&\multirow{3}{*}{0.77}&\multirow{3}{*}{-}&\multirow{3}{*}{-}&\multirow{3}{*}{-}& & 0.68$\pm$0.02&0.68$\pm$0.02&0.8$\pm$0.06&0.72$\pm$0.03&0.78$\pm$0.04&0.76$\pm$0.07&0.72$\pm$0.07&0.72$\pm$0.03&0.58$\pm$0.05&0.77$\pm$0.05 \\
 & & & & & & & (+1 noise) &0.35$\pm$0.01&0.35$\pm$0.01&0.74$\pm$0.06&0.55$\pm$0.04&0.68$\pm$0.06&0.69$\pm$0.09&0.61$\pm$0.08&0.56$\pm$0.25&0.64$\pm$0.1&0.71$\pm$0.05\\
 & & & & & & & (+2 noise) &0.34$\pm$0.0&0.34$\pm$0.0&0.72$\pm$0.04&0.47$\pm$0.02&0.62$\pm$0.05&0.79$\pm$0.05&0.72$\pm$0.07&0.55$\pm$0.27&0.7$\pm$0.1&0.65$\pm$0.05\\
 \inrule
\multirow{3}{*}{BBC}&\multirow{3}{*}{0.81}&\multirow{3}{*}{0.80}&\multirow{3}{*}{0.77}&\multirow{3}{*}{0.81}&\multirow{3}{*}{-}&\multirow{3}{*}{-}& & 0.37$\pm$0.02&0.37$\pm$0.02&0.88$\pm$0.02&0.85$\pm$0.01&0.87$\pm$0.02&0.84$\pm$0.02&0.83$\pm$0.03&0.73$\pm$0.01&0.59$\pm$0.07&0.81$\pm$0.03 \\ 
 & & & & & & & (+1 noise) &0.33$\pm$0.0&0.33$\pm$0.0&0.85$\pm$0.02&0.78$\pm$0.02&0.82$\pm$0.02&0.82$\pm$0.01&0.8$\pm$0.03&0.57$\pm$0.28&0.58$\pm$0.06&0.74$\pm$0.02\\
  & & & & & & & (+2 noise) & 0.33$\pm$0.0&0.33$\pm$0.0&0.81$\pm$0.02&0.71$\pm$0.02&0.76$\pm$0.02&0.78$\pm$0.08&0.79$\pm$0.01&0.42$\pm$0.26&0.64$\pm$0.04&0.7$\pm$0.02\\
\inrule
\multirow{3}{*}{BBCSport}&\multirow{3}{*}{0.89}&\multirow{3}{*}{0.86}&\multirow{3}{*}{-}&\multirow{3}{*}{-}&\multirow{3}{*}{-}&\multirow{3}{*}{-}& & 0.78$\pm$0.01&0.78$\pm$0.01&0.91$\pm$0.01&0.9$\pm$0.01&0.91$\pm$0.01&0.91$\pm$0.02&0.88$\pm$0.01&0.81$\pm$0.04&0.79$\pm$0.09&0.84$\pm$0.02 \\
 & & & & & & & (+1 noise) &0.36$\pm$0.0&0.36$\pm$0.0&0.82$\pm$0.03&0.79$\pm$0.03&0.8$\pm$0.02&0.87$\pm$0.01&0.87$\pm$0.01&0.41$\pm$0.28&0.75$\pm$0.11&0.68$\pm$0.04\\
  & & & & & & & (+2 noise) & 0.35$\pm$0.0&0.35$\pm$0.0&0.73$\pm$0.04&0.67$\pm$0.03&0.7$\pm$0.03&0.85$\pm$0.03&0.87$\pm$0.01&0.41$\pm$0.28&0.72$\pm$0.08&0.62$\pm$0.03\\
\inrule
\multirow{3}{*}{Wikipedia}&\multirow{3}{*}{0.17}&\multirow{3}{*}{0.60}&\multirow{3}{*}{-}&\multirow{3}{*}{-}&\multirow{3}{*}{-}&\multirow{3}{*}{-}& & 0.18$\pm$0.01&0.18$\pm$0.01&0.44$\pm$0.03&0.44$\pm$0.04&0.44$\pm$0.04&0.57$\pm$0.02&0.57$\pm$0.08&0.59$\pm$0.02&0.31$\pm$0.06&0.48$\pm$0.04 \\
 & & & & & & & (+1 noise) &0.15$\pm$0.0&0.15$\pm$0.0&0.34$\pm$0.01&0.34$\pm$0.01&0.34$\pm$0.01&0.47$\pm$0.02&0.5$\pm$0.1&0.16$\pm$0.01&0.28$\pm$0.05&0.37$\pm$0.01\\
  & & & & & & & (+2 noise) &0.15$\pm$0.0&0.15$\pm$0.0&0.3$\pm$0.01&0.29$\pm$0.01&0.3$\pm$0.01&0.42$\pm$0.03&0.57$\pm$0.02&0.16$\pm$0.01&0.23$\pm$0.0&0.32$\pm$0.02\\
\inrule
\multirow{3}{*}{UCI}&\multirow{3}{*}{0.9}&\multirow{3}{*}{0.8}&\multirow{3}{*}{0.95}&\multirow{3}{*}{0.51}&\multirow{3}{*}{0.96}&\multirow{3}{*}{0.8}& & 0.1$\pm$0.0&0.1$\pm$0.0&0.92$\pm$0.01&0.86$\pm$0.01&0.89$\pm$0.01&0.95$\pm$0.03&0.94$\pm$0.01&0.87$\pm$0.07&0.29$\pm$0.09&0.85$\pm$0.01 \\
 & & & & & &  & (+1 noise) &0.1$\pm$0.0&0.1$\pm$0.0&0.93$\pm$0.01&0.85$\pm$0.02&0.9$\pm$0.01&0.96$\pm$0.01&0.91$\pm$0.04&0.91$\pm$0.0&0.29$\pm$0.09&0.82$\pm$0.01\\
  & & & & & & & (+2 noise) & 0.1$\pm$0.0&0.1$\pm$0.0&0.93$\pm$0.01&0.82$\pm$0.02&0.89$\pm$0.01&0.96$\pm$0.01&0.94$\pm$0.01&0.91$\pm$0.0&0.26$\pm$0.06&0.81$\pm$0.02\\
 \inrule
\multirow{3}{*}{cora}&\multirow{3}{*}{0.69}&\multirow{3}{*}{0.59}&\multirow{3}{*}{-}&\multirow{3}{*}{-}&\multirow{3}{*}{-}&\multirow{3}{*}{-}& & 0.32$\pm$0.01&0.32$\pm$0.01&0.66$\pm$0.01&0.66$\pm$0.01&0.66$\pm$0.01&0.7$\pm$0.03&0.7$\pm$0.05&0.67$\pm$0.02&0.37$\pm$0.06&0.61$\pm$0.01 \\ 
 & & & & & & &(+1 noise)&0.3$\pm$0.0&0.3$\pm$0.0&0.53$\pm$0.01&0.52$\pm$0.01&0.53$\pm$0.01&0.59$\pm$0.02&0.64$\pm$0.02&0.67$\pm$0.02&0.31$\pm$0.04&0.5$\pm$0.01\\
  & & & & & & & (+2 noise) & 0.3$\pm$0.0&0.3$\pm$0.0&0.44$\pm$0.01&0.43$\pm$0.01&0.44$\pm$0.01&0.64$\pm$0.02&0.64$\pm$0.05&0.67$\pm$0.02&0.35$\pm$0.06&0.45$\pm$0.0\\
 \inrule
\multirow{3}{*}{citeseer}&\multirow{3}{*}{0.46}&\multirow{3}{*}{0.59}&\multirow{3}{*}{-}&\multirow{3}{*}{-}&\multirow{3}{*}{-}&\multirow{3}{*}{-}& & 0.27$\pm$0.0&0.27$\pm$0.0&0.62$\pm$0.01&0.62$\pm$0.01&0.62$\pm$0.01&0.63$\pm$0.01&0.63$\pm$0.01&0.45$\pm$0.02&0.26$\pm$0.05&0.51$\pm$0.01 \\ 
 & & & & & & &(+1 noise)&0.21$\pm$0.0&0.21$\pm$0.0&0.49$\pm$0.01&0.47$\pm$0.01&0.48$\pm$0.01&0.5$\pm$0.02&0.6$\pm$0.02&0.46$\pm$0.02&0.26$\pm$0.05&0.41$\pm$0.01\\
  & & & & & & & (+2 noise) &0.21$\pm$0.0&0.21$\pm$0.0&0.41$\pm$0.01&0.39$\pm$0.01&0.4$\pm$0.02&0.48$\pm$0.01&0.6$\pm$0.02&0.46$\pm$0.02&0.24$\pm$0.05&0.38$\pm$0.01\\
 \inrule
\multirow{3}{*}{dkpol}&\multirow{3}{*}{0.3}&\multirow{3}{*}{0.16}&\multirow{3}{*}{0.69}&\multirow{3}{*}{-}&\multirow{3}{*}{-}&\multirow{3}{*}{-}& & 0.15$\pm$0.0&0.15$\pm$0.0&0.7$\pm$0.04&0.69$\pm$0.03&0.7$\pm$0.03&0.65$\pm$0.05&0.59$\pm$0.09&0.27$\pm$0.03&0.24$\pm$0.08&0.59$\pm$0.04 \\ 
 & & & & & & &(+1 noise)&0.14$\pm$0.0&0.14$\pm$0.0&0.58$\pm$0.04&0.57$\pm$0.04&0.58$\pm$0.03&0.63$\pm$0.08&0.6$\pm$0.02&0.3$\pm$0.01&0.24$\pm$0.03&0.34$\pm$0.03\\
  & & & & & & & (+2 noise) &0.14$\pm$0.0&0.14$\pm$0.0&0.53$\pm$0.03&0.51$\pm$0.03&0.52$\pm$0.03&0.51$\pm$0.1&0.57$\pm$0.03&0.3$\pm$0.01&0.19$\pm$0.04&0.27$\pm$0.03\\
 \inrule
\multirow{3}{*}{aucs}&\multirow{3}{*}{0.34}&\multirow{3}{*}{0.36}&\multirow{3}{*}{0.61}&\multirow{3}{*}{0.8}&\multirow{3}{*}{0.72}&\multirow{3}{*}{-}& & 0.3$\pm$0.01&0.3$\pm$0.01&0.85$\pm$0.04&0.8$\pm$0.06&0.85$\pm$0.05&0.86$\pm$0.05&0.86$\pm$0.05&0.75$\pm$0.07&0.5$\pm$0.16&0.81$\pm$0.05 \\ 
 & & & & & & &(+1 noise)&0.27$\pm$0.0&0.27$\pm$0.0&0.81$\pm$0.04&0.37$\pm$0.03&0.65$\pm$0.03&0.81$\pm$0.04&0.81$\pm$0.04&0.76$\pm$0.09&0.48$\pm$0.08&0.77$\pm$0.09\\
  & & & & & & & (+2 noise) & 0.27$\pm$0.0&0.27$\pm$0.0&0.77$\pm$0.05&0.33$\pm$0.02&0.52$\pm$0.04&0.77$\pm$0.05&0.77$\pm$0.05&0.76$\pm$0.09&0.48$\pm$0.16&0.75$\pm$0.09\\
 \midrule
 APR & & & & & & & 
 &0.37&0.37&0.9&0.8&0.86&0.95&\textbf{0.96}&0.74&0.56&0.8\\
  AR & & & & & & & 
 & 9.3&9.3&3.0&5.4&4.2&\textbf{2.3}& 2.8&5.6&7.6&5.6\\
\bottomrule
\end{NiceTabular}}
\end{scriptsize}
\end{table*}
 \begin{table*}[h!]
\caption{Accuracy (mean $\pm$ std) over three random samples of the 10\% of input labels, on real-world datasets (+ one and two layers of noise).}
\label{tab:2layers10percent}
\centering
\begin{scriptsize}
\resizebox{\textwidth}{!}{
\begin{NiceTabular}{c|cccccc|cccccc|cc|ccc}
\toprule
 & \RNum{1} & \RNum{2} & \RNum{3} & \RNum{4} & \RNum{5} & \RNum{6} &  & MIN & GEOM & ARIT & HARM & MAX & BINOM & MULTI & SGMI & SMACD & GMM\\
\midrule
\multirow{3}{*}{3sources}&\multirow{3}{*}{0.76}&\multirow{3}{*}{0.72}&\multirow{3}{*}{0.75}&\multirow{3}{*}{-}&\multirow{3}{*}{-}&\multirow{3}{*}{-}& & 0.71$\pm$0.07&0.71$\pm$0.07&0.79$\pm$0.07&0.69$\pm$0.05&0.75$\pm$0.05&0.75$\pm$0.07&0.74$\pm$0.05&0.75$\pm$0.04&0.62$\pm$0.04&0.8$\pm$0.05 \\
 & & & & & & & (+1 noise) &0.34$\pm$0.01&0.34$\pm$0.01&0.74$\pm$0.1&0.54$\pm$0.06&0.66$\pm$0.09&0.76$\pm$0.11&0.73$\pm$0.06&0.58$\pm$0.27&0.6$\pm$0.08&0.76$\pm$0.04\\
 & & & & & & & (+2 noise) &0.33$\pm$0.0&0.33$\pm$0.0&0.7$\pm$0.08&0.46$\pm$0.06&0.6$\pm$0.08&0.72$\pm$0.15&0.75$\pm$0.07&0.57$\pm$0.28&0.61$\pm$0.21&0.71$\pm$0.05\\
 \inrule
\multirow{3}{*}{BBC}&\multirow{3}{*}{0.83}&\multirow{3}{*}{0.83}&\multirow{3}{*}{0.79}&\multirow{3}{*}{0.83}&\multirow{3}{*}{-}&\multirow{3}{*}{-}& & 0.38$\pm$0.01&0.38$\pm$0.01&0.91$\pm$0.01&0.89$\pm$0.01&0.9$\pm$0.01&0.88$\pm$0.01&0.86$\pm$0.04&0.76$\pm$0.02&0.69$\pm$0.03&0.87$\pm$0.01 \\ 
 & & & & & & & (+1 noise) &0.33$\pm$0.0&0.33$\pm$0.0&0.89$\pm$0.01&0.84$\pm$0.01&0.87$\pm$0.02&0.87$\pm$0.02&0.87$\pm$0.01&0.76$\pm$0.03&0.66$\pm$0.02&0.84$\pm$0.01\\
  & & & & & & & (+2 noise) & 0.33$\pm$0.0&0.33$\pm$0.0&0.88$\pm$0.01&0.79$\pm$0.01&0.84$\pm$0.02&0.87$\pm$0.01&0.83$\pm$0.03&0.76$\pm$0.03&0.65$\pm$0.07&0.81$\pm$0.01\\
\inrule
\multirow{3}{*}{BBCSport}&\multirow{3}{*}{0.90}&\multirow{3}{*}{0.88}&\multirow{3}{*}{-}&\multirow{3}{*}{-}&\multirow{3}{*}{-}&\multirow{3}{*}{-}& & 0.79$\pm$0.0&0.79$\pm$0.0&0.92$\pm$0.01&0.92$\pm$0.0&0.92$\pm$0.0&0.92$\pm$0.02&0.88$\pm$0.01&0.84$\pm$0.03&0.73$\pm$0.1&0.88$\pm$0.02 \\
 & & & & & & & (+1 noise) &0.36$\pm$0.0&0.36$\pm$0.0&0.86$\pm$0.02&0.82$\pm$0.01&0.85$\pm$0.02&0.91$\pm$0.01&0.87$\pm$0.0&0.83$\pm$0.04&0.6$\pm$0.07&0.77$\pm$0.02\\
  & & & & & & & (+2 noise) & 0.35$\pm$0.0&0.35$\pm$0.0&0.8$\pm$0.02&0.74$\pm$0.02&0.77$\pm$0.02&0.9$\pm$0.01&0.87$\pm$0.01&0.64$\pm$0.31&0.78$\pm$0.03&0.73$\pm$0.03\\
\inrule
\multirow{3}{*}{Wikipedia}&\multirow{3}{*}{0.18}&\multirow{3}{*}{0.65}&\multirow{3}{*}{-}&\multirow{3}{*}{-}&\multirow{3}{*}{-}&\multirow{3}{*}{-}& & 0.19$\pm$0.01&0.19$\pm$0.01&0.51$\pm$0.03&0.51$\pm$0.03&0.51$\pm$0.03&0.62$\pm$0.04&0.64$\pm$0.02&0.61$\pm$0.03&0.24$\pm$0.04&0.57$\pm$0.03 \\
 & & & & & & & (+1 noise) &0.15$\pm$0.0&0.15$\pm$0.0&0.42$\pm$0.01&0.42$\pm$0.01&0.42$\pm$0.01&0.57$\pm$0.06&0.62$\pm$0.03&0.59$\pm$0.02&0.25$\pm$0.04&0.47$\pm$0.02\\
  & & & & & & & (+2 noise) &0.15$\pm$0.0&0.15$\pm$0.0&0.39$\pm$0.01&0.38$\pm$0.01&0.38$\pm$0.01&0.56$\pm$0.06&0.59$\pm$0.03&0.59$\pm$0.02&0.23$\pm$0.08&0.43$\pm$0.02\\
\inrule
\multirow{3}{*}{UCI}&\multirow{3}{*}{0.92}&\multirow{3}{*}{0.81}&\multirow{3}{*}{0.96}&\multirow{3}{*}{0.57}&\multirow{3}{*}{0.97}&\multirow{3}{*}{0.82}& & 0.11$\pm$0.0&0.11$\pm$0.0&0.95$\pm$0.01&0.88$\pm$0.01&0.92$\pm$0.01&0.97$\pm$0.01&0.96$\pm$0.01&0.94$\pm$0.01&0.33$\pm$0.05&0.93$\pm$0.01 \\
 & & & & & &  & (+1 noise) &0.1$\pm$0.0&0.1$\pm$0.0&0.96$\pm$0.0&0.88$\pm$0.01&0.92$\pm$0.0&0.97$\pm$0.0&0.96$\pm$0.01&0.94$\pm$0.01&0.3$\pm$0.1&0.93$\pm$0.01\\
  & & & & & & & (+2 noise) & 0.1$\pm$0.0&0.1$\pm$0.0&0.96$\pm$0.0&0.88$\pm$0.01&0.92$\pm$0.0&0.97$\pm$0.0&0.96$\pm$0.01&0.94$\pm$0.01&0.33$\pm$0.05&0.93$\pm$0.01\\
 \inrule
\multirow{3}{*}{cora}&\multirow{3}{*}{0.73}&\multirow{3}{*}{0.62}&\multirow{3}{*}{-}&\multirow{3}{*}{-}&\multirow{3}{*}{-}&\multirow{3}{*}{-}& & 0.34$\pm$0.01&0.34$\pm$0.01&0.69$\pm$0.01&0.69$\pm$0.01&0.69$\pm$0.01&0.74$\pm$0.03&0.76$\pm$0.01&0.72$\pm$0.01&0.34$\pm$0.08&0.69$\pm$0.01 \\ 
 & & & & & & &(+1 noise)&0.3$\pm$0.0&0.3$\pm$0.0&0.57$\pm$0.01&0.56$\pm$0.01&0.57$\pm$0.01&0.63$\pm$0.01&0.76$\pm$0.01&0.72$\pm$0.02&0.28$\pm$0.01&0.58$\pm$0.01\\
  & & & & & & & (+2 noise) & 0.3$\pm$0.0&0.3$\pm$0.0&0.49$\pm$0.01&0.47$\pm$0.01&0.48$\pm$0.01&0.64$\pm$0.04&0.74$\pm$0.03&0.72$\pm$0.02&0.37$\pm$0.02&0.55$\pm$0.0\\
 \inrule
\multirow{3}{*}{citeseer}&\multirow{3}{*}{0.51}&\multirow{3}{*}{0.62}&\multirow{3}{*}{-}&\multirow{3}{*}{-}&\multirow{3}{*}{-}&\multirow{3}{*}{-}& & 0.29$\pm$0.01&0.29$\pm$0.01&0.65$\pm$0.01&0.65$\pm$0.01&0.65$\pm$0.01&0.66$\pm$0.01&0.65$\pm$0.01&0.51$\pm$0.01&0.36$\pm$0.1&0.58$\pm$0.01 \\ 
 & & & & & & &(+1 noise)&0.21$\pm$0.0&0.21$\pm$0.0&0.53$\pm$0.02&0.52$\pm$0.02&0.53$\pm$0.02&0.59$\pm$0.03&0.63$\pm$0.0&0.52$\pm$0.02&0.32$\pm$0.03&0.49$\pm$0.01\\
  & & & & & & & (+2 noise) &0.21$\pm$0.0&0.21$\pm$0.0&0.47$\pm$0.02&0.44$\pm$0.01&0.46$\pm$0.01&0.61$\pm$0.03&0.63$\pm$0.0&0.52$\pm$0.02&0.26$\pm$0.05&0.45$\pm$0.01\\
 \inrule
\multirow{3}{*}{dkpol}&\multirow{3}{*}{0.32}&\multirow{3}{*}{0.16}&\multirow{3}{*}{0.69}&\multirow{3}{*}{-}&\multirow{3}{*}{-}&\multirow{3}{*}{-}& & 0.16$\pm$0.0&0.16$\pm$0.0&0.73$\pm$0.06&0.67$\pm$0.09&0.69$\pm$0.07&0.62$\pm$0.06&0.76$\pm$0.01&0.31$\pm$0.03&0.26$\pm$0.09&0.63$\pm$0.04 \\ 
 & & & & & & &(+1 noise)&0.14$\pm$0.0&0.14$\pm$0.0&0.62$\pm$0.02&0.59$\pm$0.02&0.6$\pm$0.02&0.54$\pm$0.08&0.72$\pm$0.08&0.31$\pm$0.05&0.24$\pm$0.01&0.34$\pm$0.01\\
  & & & & & & & (+2 noise) &0.14$\pm$0.0&0.14$\pm$0.0&0.58$\pm$0.02&0.54$\pm$0.03&0.55$\pm$0.02&0.45$\pm$0.11&0.69$\pm$0.07&0.31$\pm$0.05&0.2$\pm$0.04&0.27$\pm$0.02\\
 \inrule
\multirow{3}{*}{aucs}&\multirow{3}{*}{0.34}&\multirow{3}{*}{0.36}&\multirow{3}{*}{0.61}&\multirow{3}{*}{0.8}&\multirow{3}{*}{0.72}&\multirow{3}{*}{-}& & 0.3$\pm$0.01&0.3$\pm$0.01&0.85$\pm$0.04&0.8$\pm$0.06&0.85$\pm$0.05&0.85$\pm$0.04&0.85$\pm$0.02&0.75$\pm$0.07&0.58$\pm$0.03&0.81$\pm$0.05 \\ 
 & & & & & & &(+1 noise)&0.27$\pm$0.0&0.27$\pm$0.0&0.81$\pm$0.04&0.37$\pm$0.03&0.65$\pm$0.03&0.81$\pm$0.04&0.81$\pm$0.04&0.76$\pm$0.09&0.56$\pm$0.07&0.77$\pm$0.09\\
  & & & & & & & (+2 noise) & 0.27$\pm$0.0&0.27$\pm$0.0&0.77$\pm$0.05&0.33$\pm$0.02&0.52$\pm$0.04&0.77$\pm$0.05&0.77$\pm$0.05&0.76$\pm$0.09&0.6$\pm$0.09&0.76$\pm$0.07\\
 \midrule
 APR & & & & & & & 
 &0.36&0.36&0.9&0.8&0.86&0.94&\textbf{0.98}&0.84&0.56&0.85\\
  AR & & & & & & & 
 & 9.3&9.3&3.2&6.1&4.7&2.5& \textbf{2.4}&4.8&7.8&5.0\\
\bottomrule
\end{NiceTabular}}
\end{scriptsize}
\end{table*}
 \begin{table*}[h!]
\caption{Accuracy (mean $\pm$ std) over three random samples of the 15\% of input labels, on real-world datasets (+ one and two layers of noise).}
\label{tab:2layers15percent}
\centering
\begin{scriptsize}
\resizebox{\textwidth}{!}{
\begin{NiceTabular}{c|cccccc|cccccc|cc|ccc}
\toprule
 & \RNum{1} & \RNum{2} & \RNum{3} & \RNum{4} & \RNum{5} & \RNum{6} &  & MIN & GEOM & ARIT & HARM & MAX & BINOM & MULTI & SGMI & SMACD & GMM\\
\midrule
\multirow{3}{*}{3sources}&\multirow{3}{*}{0.77}&\multirow{3}{*}{0.75}&\multirow{3}{*}{0.78}&\multirow{3}{*}{-}&\multirow{3}{*}{-}&\multirow{3}{*}{-}& & 0.74$\pm$0.06 & 0.74$\pm$0.06 & 0.83$\pm$0.05 & 0.75$\pm$0.04 & 0.8$\pm$0.03 & 0.82$\pm$0.03 & 0.76$\pm$0.04 & 0.74$\pm$0.04 & 0.66$\pm$0.09 & 0.8$\pm$0.03 \\
 & & & & & & & (+1 noise) &0.34$\pm$0.01 & 0.34$\pm$0.01 & 0.79$\pm$0.05 & 0.57$\pm$0.04 & 0.71$\pm$0.03 & 0.79$\pm$0.06 & 0.72$\pm$0.07 & 0.75$\pm$0.05 & 0.61$\pm$0.09 & 0.77$\pm$0.02\\
 & & & & & & & (+2 noise) &0.34$\pm$0.0&0.34$\pm$0.0&0.76$\pm$0.09&0.47$\pm$0.04&0.63$\pm$0.04&0.75$\pm$0.08&0.72$\pm$0.02&0.75$\pm$0.05&0.63$\pm$0.15&0.76$\pm$0.07\\
 \inrule
\multirow{3}{*}{BBC}&\multirow{3}{*}{0.85}&\multirow{3}{*}{0.84}&\multirow{3}{*}{0.8}&\multirow{3}{*}{0.84}&\multirow{3}{*}{-}&\multirow{3}{*}{-}& & 0.39$\pm$0.02 & 0.39$\pm$0.02 & 0.91$\pm$0.01 & 0.89$\pm$0.01 & 0.9$\pm$0.01 & 0.89$\pm$0.01 & 0.85$\pm$0.03 & 0.77$\pm$0.02 & 0.64$\pm$0.12 & 0.89$\pm$0.01 \\ 
 & & & & & & & (+1 noise) &0.33$\pm$0.0 & 0.33$\pm$0.0 & 0.9$\pm$0.01 & 0.86$\pm$0.0 & 0.88$\pm$0.01 & 0.89$\pm$0.01 & 0.86$\pm$0.03 & 0.79$\pm$0.01 & 0.62$\pm$0.05 & 0.87$\pm$0.01\\
  & & & & & & & (+2 noise) & 0.33$\pm$0.0&0.33$\pm$0.0&0.88$\pm$0.01&0.83$\pm$0.01&0.86$\pm$0.0&0.88$\pm$0.01&0.85$\pm$0.03&0.79$\pm$0.01&0.64$\pm$0.01&0.84$\pm$0.01\\
\inrule
\multirow{3}{*}{BBCSport}&\multirow{3}{*}{0.91}&\multirow{3}{*}{0.9}&\multirow{3}{*}{-}&\multirow{3}{*}{-}&\multirow{3}{*}{-}&\multirow{3}{*}{-}& & 0.81$\pm$0.01 & 0.81$\pm$0.01 & 0.94$\pm$0.01 & 0.93$\pm$0.0 & 0.93$\pm$0.0 & 0.94$\pm$0.01 & 0.91$\pm$0.02 & 0.84$\pm$0.02 & 0.77$\pm$0.09 & 0.91$\pm$0.02 \\
 & & & & & & & (+1 noise) &0.36$\pm$0.0 & 0.36$\pm$0.0 & 0.9$\pm$0.01 & 0.87$\pm$0.01 & 0.89$\pm$0.0 & 0.91$\pm$0.02 & 0.87$\pm$0.02 & 0.84$\pm$0.03 & 0.85$\pm$0.02 & 0.86$\pm$0.02\\
  & & & & & & & (+2 noise) & 0.35$\pm$0.0&0.35$\pm$0.0&0.87$\pm$0.01&0.81$\pm$0.01&0.84$\pm$0.01&0.94$\pm$0.01&0.88$\pm$0.01&0.84$\pm$0.03&0.85$\pm$0.07&0.8$\pm$0.03\\
\inrule
\multirow{3}{*}{Wikipedia}&\multirow{3}{*}{0.17}&\multirow{3}{*}{0.65}&\multirow{3}{*}{-}&\multirow{3}{*}{-}&\multirow{3}{*}{-}&\multirow{3}{*}{-}& & 0.21$\pm$0.01 & 0.21$\pm$0.01 & 0.56$\pm$0.02 & 0.56$\pm$0.02 & 0.56$\pm$0.02 & 0.64$\pm$0.01 & 0.64$\pm$0.02 & 0.62$\pm$0.02 & 0.31$\pm$0.05 & 0.62$\pm$0.01 \\
 & & & & & & & (+1 noise) &0.15$\pm$0.0 & 0.15$\pm$0.0 & 0.51$\pm$0.01 & 0.51$\pm$0.01 & 0.52$\pm$0.01 & 0.61$\pm$0.02 & 0.63$\pm$0.03 & 0.61$\pm$0.01 & 0.32$\pm$0.06 & 0.55$\pm$0.01\\
  & & & & & & & (+2 noise) &0.15$\pm$0.0&0.15$\pm$0.0&0.48$\pm$0.01&0.46$\pm$0.03&0.47$\pm$0.02&0.6$\pm$0.02&0.66$\pm$0.01&0.61$\pm$0.01&0.23$\pm$0.02&0.51$\pm$0.01\\
\inrule
\multirow{3}{*}{UCI}&\multirow{3}{*}{0.92}&\multirow{3}{*}{0.82}&\multirow{3}{*}{0.96}&\multirow{3}{*}{0.59}&\multirow{3}{*}{0.97}&\multirow{3}{*}{0.83}& & 0.11$\pm$0.0 & 0.11$\pm$0.0 & 0.96$\pm$0.01 & 0.89$\pm$0.01 & 0.93$\pm$0.0 & 0.97$\pm$0.0 & 0.97$\pm$0.01 & 0.95$\pm$0.01 & 0.35$\pm$0.11 & 0.96$\pm$0.01 \\
 & & & & & &  & (+1 noise) &0.1$\pm$0.0 & 0.1$\pm$0.0 & 0.97$\pm$0.0 & 0.9$\pm$0.0 & 0.94$\pm$0.0 & 0.97$\pm$0.0 & 0.97$\pm$0.01 & 0.95$\pm$0.01 & 0.26$\pm$0.06 & 0.96$\pm$0.01\\
  & & & & & & & (+2 noise) & 0.1$\pm$0.0&0.1$\pm$0.0&0.97$\pm$0.0&0.89$\pm$0.01&0.94$\pm$0.0&0.96$\pm$0.01&0.97$\pm$0.01&0.95$\pm$0.01&0.29$\pm$0.01&0.96$\pm$0.01\\
 \inrule
\multirow{3}{*}{cora}&\multirow{3}{*}{0.75}&\multirow{3}{*}{0.64}&\multirow{3}{*}{-}&\multirow{3}{*}{-}&\multirow{3}{*}{-}&\multirow{3}{*}{-}& & 0.35$\pm$0.01 & 0.35$\pm$0.01 & 0.71$\pm$0.0 & 0.7$\pm$0.0 & 0.71$\pm$0.0 & 0.77$\pm$0.04 & 0.77$\pm$0.02 & 0.75$\pm$0.01 & 0.41$\pm$0.03 & 0.73$\pm$0.01 \\ 
 & & & & & & &(+1 noise)&0.3$\pm$0.0 & 0.3$\pm$0.0 & 0.61$\pm$0.0 & 0.59$\pm$0.0 & 0.6$\pm$0.0 & 0.65$\pm$0.02 & 0.66$\pm$0.07 & 0.75$\pm$0.01 & 0.35$\pm$0.07 & 0.65$\pm$0.0\\
  & & & & & & & (+2 noise) & 0.3$\pm$0.0&0.3$\pm$0.0&0.54$\pm$0.01&0.52$\pm$0.01&0.53$\pm$0.0&0.7$\pm$0.03&0.72$\pm$0.01&0.75$\pm$0.01&0.36$\pm$0.05&0.62$\pm$0.01\\
 \inrule
\multirow{3}{*}{citeseer}&\multirow{3}{*}{0.56}&\multirow{3}{*}{0.65}&\multirow{3}{*}{-}&\multirow{3}{*}{-}&\multirow{3}{*}{-}&\multirow{3}{*}{-}& & 0.31$\pm$0.01 & 0.31$\pm$0.01 & 0.67$\pm$0.0 & 0.67$\pm$0.01 & 0.67$\pm$0.01 & 0.68$\pm$0.01 & 0.68$\pm$0.01 & 0.55$\pm$0.02 & 0.39$\pm$0.08 & 0.61$\pm$0.01 \\ 
 & & & & & & &(+1 noise)&0.21$\pm$0.0 & 0.21$\pm$0.0 & 0.57$\pm$0.0 & 0.56$\pm$0.0 & 0.57$\pm$0.01 & 0.62$\pm$0.03 & 0.66$\pm$0.03 & 0.55$\pm$0.03 & 0.35$\pm$0.07 & 0.54$\pm$0.01\\
  & & & & & & & (+2 noise) &0.21$\pm$0.0&0.21$\pm$0.0&0.51$\pm$0.01&0.48$\pm$0.01&0.5$\pm$0.01&0.66$\pm$0.04&0.64$\pm$0.01&0.55$\pm$0.03&0.36$\pm$0.13&0.52$\pm$0.01\\
 \inrule
\multirow{3}{*}{dkpol}&\multirow{3}{*}{0.35}&\multirow{3}{*}{0.16}&\multirow{3}{*}{0.69}&\multirow{3}{*}{-}&\multirow{3}{*}{-}&\multirow{3}{*}{-}& & 0.16$\pm$0.01 & 0.16$\pm$0.01 & 0.75$\pm$0.05 & 0.68$\pm$0.06 & 0.7$\pm$0.05 & 0.67$\pm$0.05 & 0.8$\pm$0.04 & 0.34$\pm$0.04 & 0.25$\pm$0.06 & 0.69$\pm$0.05 \\ 
 & & & & & & &(+1 noise)&0.14$\pm$0.0 & 0.14$\pm$0.0 & 0.65$\pm$0.04 & 0.61$\pm$0.02 & 0.63$\pm$0.03 & 0.5$\pm$0.06 & 0.71$\pm$0.09 & 0.36$\pm$0.05 & 0.26$\pm$0.05 & 0.39$\pm$0.04\\
  & & & & & & & (+2 noise) &0.14$\pm$0.0&0.14$\pm$0.0&0.6$\pm$0.02&0.56$\pm$0.01&0.57$\pm$0.02&0.52$\pm$0.07&0.73$\pm$0.07&0.36$\pm$0.05&0.19$\pm$0.04&0.32$\pm$0.05\\
 \inrule
\multirow{3}{*}{aucs}&\multirow{3}{*}{0.34}&\multirow{3}{*}{0.36}&\multirow{3}{*}{0.61}&\multirow{3}{*}{0.8}&\multirow{3}{*}{0.72}&\multirow{3}{*}{-}& & 0.3$\pm$0.01 & 0.3$\pm$0.01 & 0.85$\pm$0.04 & 0.8$\pm$0.06 & 0.85$\pm$0.05 & 0.84$\pm$0.04 & 0.85$\pm$0.04 & 0.75$\pm$0.07 & 0.55$\pm$0.04 & 0.81$\pm$0.05 \\ 
 & & & & & & &(+1 noise)&0.28$\pm$0.01 & 0.28$\pm$0.01 & 0.82$\pm$0.05 & 0.44$\pm$0.02 & 0.73$\pm$0.04 & 0.82$\pm$0.05 & 0.82$\pm$0.05 & 0.76$\pm$0.09 & 0.48$\pm$0.06 & 0.78$\pm$0.05\\
  & & & & & & & (+2 noise) & 0.27$\pm$0.0&0.27$\pm$0.0&0.78$\pm$0.04&0.32$\pm$0.05&0.57$\pm$0.03&0.78$\pm$0.04&0.78$\pm$0.04&0.76$\pm$0.09&0.54$\pm$0.05&0.78$\pm$0.07\\
 \midrule
 APR & & & & & & & 
 &0.36&0.36&0.93&0.83&0.89&0.95&\textbf{0.97}&0.87&0.57&0.89\\
  AR & & & & & & & 
 & 9.4&9.4&3.1&6.1&4.7&\textbf{2.6}& 2.7&4.9&7.8&4.3\\
\bottomrule
\end{NiceTabular}}
\end{scriptsize}
\end{table*}

\clearpage
\section{Learned Parameters} \label{AppendixF}
In Table~\ref{tab:Learned parameters}, we report the different parameters learned by the methods on the real datasets in Subsection~\ref{Real World Datasets}. The numbers are averaged over three random samplings of the initially labeled nodes. We emphasize that:
\begin{itemize}
    \item SGMI always assigns all the weight to one single layer, which may change when the initial labels change;
    \item SMACD computes a coupled matrix-tensor nonnegative factorization. The parameters shown here are the norm of the rows of the coupling kernel in the computed factorization (averaged over the 3 runs);
    \item GMM has only one parameter, and it is always fixed a-priori to $p=-1$;
    \item BINOM learns different parameters for different classes (which are denoted as $B_1$ $B_2$ $B_3$ and so forth).
\end{itemize}

\setlength{\tabcolsep}{2.5pt}
 \begin{table*}[h!]
\caption{Different parameters learned by the methods on real datasets. The numbers are averaged over three random samplings of the initially labeled nodes.}
\label{tab:Learned parameters}
\vskip 0.15in
\centering
\begin{scriptsize}
    \begin{NiceTabular}{lcccccccccccccc}
    \toprule
        Dataset & ~ & $B_1$ & $B_2$ & $B_3$ & $B_4$ & $B_5$ & $B_6$ & $B_7$ & $B_8$ & $B_9$ & $B_{10}$ & MULTI & SGMI & SMACD \\ 
        \midrule
        3sources &         
        $\boldsymbol{\beta}_1$ & 0.39 & 0.59 & 0.33 & 0.6 & 0.59 & 0.33 &  &  &  &  & 0.6 & 0.6 & 0.32 \\ 
        ~ & $\boldsymbol{\beta}_2$ & 0.21 & 0.19 & 0.33 & 0.2 & 0.22 & 0.33 &  &  &  &  & 0.2 & 0.4 & 0.33 \\ 
        ~ & $\boldsymbol{\beta}_3$ & 0.39 & 0.22 & 0.33 & 0.19 & 0.19 & 0.33 &  &  &  &  & 0.2 & 0 & 0.35 \\ 
        ~ & $\alpha$ & 0.14 & 0.14 & 1 & 8.1 & 0.17 & 1 &  &  &  &  & -17.94 &  & \\ 
        ~ & $\lambda$ & 4.82 & 4.76 & 1 & 0.89 & 4.81 & 1 &  &  &  &  & 0.1 &  &  \\
        \hline      
        BBC & 
        $\boldsymbol{\beta}_1$ & 0.23 & 0.17 & 0.26 & 0.24 & 0.27 &  &  &  &  &  & 0.16 & 0.2 & 0.25 \\ 
        ~ & $\boldsymbol{\beta}_2$ & 0.21 & 0.17 & 0.18 & 0.26 & 0.23 &  &  &  &  &  & 0.54 & 0.4 & 0.25 \\ 
        ~ & $\boldsymbol{\beta}_3$ & 0.25 & 0.12 & 0.19 & 0.15 & 0.15 &  &  &  &  &  & 0.15 & 0.2 & 0.24 \\ 
        ~ & $\boldsymbol{\beta}_4$ & 0.3 & 0.54 & 0.37 & 0.35 & 0.35 &  &  &  &  &  & 0.15 & 0.2 & 0.26 \\ 
        ~ & $\alpha$ & 2.87 & -3.69 & 0.42 & -7.88 & 0.39 &  &  &  &  &  & 0.13 &  &  \\ 
        ~ & $\lambda$ & 1.66 & 3.58 & 1.77 & 4.33 & 1.75 &  &  &  &  &  & 0.68 &  &  \\\hline        
        BBCSport & 
        $\boldsymbol{\beta}_1$ & 0.57 & 0.46 & 0.6 & 0.45 & 0.42 &  &  &  &  &  & 0.44 & 0.8 & 0.48 \\ 
        ~ & $\boldsymbol{\beta}_2$ & 0.43 & 0.54 & 0.4 & 0.55 & 0.58 &  &  &  &  &  & 0.56 & 0.2 & 0.52 \\ 
        ~ & $\alpha$ & 4.21 & 0.27 & -4 & 0.31 & 0.16 &  &  &  &  &  & 0.08 &  & \\ 
        ~ & $\lambda$ & 3.75 & 2.59 & 10 & 2.58 & 2.43 &  &  &  &  &  & 2.22 &  &   \\\hline      
        Wikipedia & $\boldsymbol{\beta}_1$ & 0.32 & 0.34 & 0.27 & 0.35 & 0.32 & 0.21 & 0.27 & 0.42 & 0.27 & 0.07 & 0.2 & 0 & 0.17 \\ 
        ~ & $\boldsymbol{\beta}_2$ & 0.68 & 0.66 & 0.73 & 0.65 & 0.68 & 0.79 & 0.73 & 0.58 & 0.73 & 0.93 & 0.8 & 1 & 0.83 \\ 
        ~ & $\alpha$ & 6.62 & 0.18 & -3.78 & 0.46 & 8.8 & -7.85 & -3.83 & 2.48 & 4.16 & 8.15 & 0.13 &  &  \\ 
        ~ & $\lambda$ & 8.38 & 4.71 & 5.3 & 4.78 & 7.23 & 6.48 & 5.27 & 5.41 & 5.9 & 8.88 & 6.45 &  &   \\\hline        
        UCI & $\boldsymbol{\beta}_1$ & 0.02 & 0.09 & 0.06 & 0.06 & 0.14 & 0.09 & 0.21 & 0.09 & 0.09 & 0 & 0.16 & 0 & 0.17 \\ 
        ~ & $\boldsymbol{\beta}_2$ & 0.43 & 0.1 & 0.27 & 0.07 & 0.11 & 0.27 & 0 & 0.08 & 0.51 & 0 & 0.16 & 0 & 0.17 \\ 
        ~ & $\boldsymbol{\beta}_3$ & 0 & 0.26 & 0.29 & 0.23 & 0.12 & 0.31 & 0.4 & 0.08 & 0.07 & 0.4 & 0.17 & 0.2 & 0.18 \\ 
        ~ & $\boldsymbol{\beta}_4$ & 0.23 & 0.09 & 0.05 & 0.06 & 0.12 & 0.09 & 0 & 0.09 & 0.09 & 0 & 0.16 & 0 & 0.13 \\ 
        ~ & $\boldsymbol{\beta}_5$ & 0.09 & 0.37 & 0.29 & 0.52 & 0.39 & 0.14 & 0.4 & 0.58 & 0.15 & 0.6 & 0.2 & 0.8 & 0.18 \\ 
        ~ & $\boldsymbol{\beta}_6$ & 0.23 & 0.09 & 0.05 & 0.06 & 0.11 & 0.09 & 0 & 0.08 & 0.09 & 0 & 0.16 & 0 & 0.17 \\ 
        ~ & $\alpha$ & 10.05 & 2.28 & 4.21 & -9.55 & 6.36 & 0.26 & -3.51 & 2.31 & 2.35 & -12 & 0.22 &  &  \\ 
        ~ & $\lambda$ & 8.04 & 4.46 & 6.52 & 4.34 & 2.79 & 4.77 & 8.19 & 4.59 & 4.47 & 10 & 1.03 &  &  \\\hline        
        cora & $\boldsymbol{\beta}_1$ & 0.73 & 0.93 & 0.75 & 0.73 & 0.83 & 0.73 & 0.74 &  &  &  & 0.71 & 1 & 0.07 \\ 
        ~ & $\boldsymbol{\beta}_2$ & 0.27 & 0.07 & 0.25 & 0.27 & 0.17 & 0.27 & 0.26 &  &  &  & 0.29 & 0 & 0.93 \\ 
        ~ & $\alpha$ & 1.05 & -7.73 & 1.95 & 0.71 & -3.6 & 0.66 & 1.45 &  &  &  & 0.28 &  &  \\ 
        ~ & $\lambda$ & 2.31 & 5.67 & 2.69 & 2.32 & 3.88 & 2.18 & 2.58 &  &  &  & 0.91 &  &   \\\hline        
        citeseer & $\boldsymbol{\beta}_1$ & 0.29 & 0.5 & 0.44 & 0.46 & 0.54 & 0.43 &  &  &  &  & 0.67 & 1 & 0.31\\ 
        ~ & $\boldsymbol{\beta}_2$ & 0.71 & 0.5 & 0.56 & 0.54 & 0.46 & 0.57 &  &  &  &  & 0.33 & 0 & 0.69 \\ 
        ~ & $\alpha$ & 8.03 & 1.87 & 3.11 & 2.17 & 3.27 & 3.24 &  &  &  &  & 0.49 &  &  \\ 
        ~ & $\lambda$ & 4.92 & 2.04 & 2.69 & 2.13 & 2.5 & 2.66 &  &  &  &  & 1.09 &  &   \\\hline        
        dkpol & $\boldsymbol{\beta}_1$ & 0.2 & 0.15 & 0.26 & 0.03 & 0.24 & 0.19 & 0.24 & 0.19 & 0.22 & 0.2 & 0.36 & 1 & 0.03 \\ 
        ~ & $\boldsymbol{\beta}_2$ & 0.24 & 0.2 & 0.23 & 0.14 & 0.35 & 0.23 & 0.35 & 0.19 & 0.24 & 0.26 & 0.32 & 0 & 0.01 \\ 
        ~ & $\boldsymbol{\beta}_3$ & 0.57 & 0.66 & 0.51 & 0.82 & 0.42 & 0.58 & 0.42 & 0.62 & 0.54 & 0.54 & 0.32 & 0 & 0.96 \\ 
        ~ & $\alpha$ & 6.35 & 8.82 & 5.29 & 12.55 & 0.78 & 6.77 & 0.62 & 9.29 & -2.21 & 5.63 & 0.26 &  &  \\ 
        ~ & $\lambda$ & 4.3 & 5.45 & 4.65 & 7.77 & 2.24 & 4.39 & 2.31 & 5.67 & 2.22 & 3.94 & 0.97 &  &   \\\hline        
        aucs & $\boldsymbol{\beta}_1$ & 0.17 & 0.17 & 0.17 & 0.17 & 0.17 & 0.17 & 0.17 & 0.17 & 0.17 &  & 0.21 & 0 & 0.03\\ 
        ~ & $\boldsymbol{\beta}_2$ & 0.19 & 0.19 & 0.19 & 0.19 & 0.19 & 0.19 & 0.19 & 0.19 & 0.19 &  & 0.15 & 0 & 0.18 \\ 
        ~ & $\boldsymbol{\beta}_3$ & 0.13 & 0.13 & 0.13 & 0.13 & 0.13 & 0.13 & 0.13 & 0.13 & 0.13 &  & 0.1 & 0 & 0.16 \\ 
        ~ & $\boldsymbol{\beta}_4$ & 0.28 & 0.28 & 0.28 & 0.28 & 0.28 & 0.28 & 0.28 & 0.28 & 0.28 &  & 0.42 & 0.8 & 0.31\\ 
        ~ & $\boldsymbol{\beta}_5$ & 0.23 & 0.23 & 0.23 & 0.23 & 0.23 & 0.23 & 0.23 & 0.23 & 0.23 &  & 0.12 & 0.2 & 0.31 \\ 
        ~ & $\alpha$ & 1.38 & 1.38 & 1.38 & 1.38 & 1.38 & 1.38 & 1.38 & 1.38 & 1.38 &  & 1.05 &  & \\
        ~ & $\lambda$ & 0.7 & 0.7 & 0.7 & 0.7 & 0.7 & 0.7 & 0.7 & 0.7 & 0.7 &  & 2.56 &  &  \\ 
        \bottomrule
    \end{NiceTabular}
    \end{scriptsize}
\end{table*}

\end{document}